\documentclass[10pt,journal,compsoc]{IEEEtran}

\usepackage[pdftex]{graphicx}

\usepackage{amsmath}
\usepackage{algorithmic}
\usepackage{amssymb}
\usepackage{multirow}
\usepackage{fixltx2e}

\begin{document}

\title{DeHIN: A Decentralized Framework for Embedding Large-scale Heterogeneous Information Networks}

\author{Mubashir~Imran, Hongzhi~Yin, Tong~Chen, Zi~Huang, Kai~Zheng\\
	\IEEEcompsocitemizethanks{
		\IEEEcompsocthanksitem M. Imran, H. Yin, T. Chen, and Z. Huang are with the School of Information Technology and Electrical Engineering, The University of Queensland, Brisbane, Australia\protect\\
		E-mail: m.imran@uq.net.au, h.yin1@uq.edu.au, tong.chen@uq.edu.au, huang@itee.uq.edu.au
		\IEEEcompsocthanksitem K. Zheng is with the School of Computer Science and Engineering, University of Electronic Science and Technology of China, Chengdu, China.\protect\\
		E-mail: zhengkai@uestc.edu.cn
}
\thanks{Hongzhi Yin is the corresponding author.}
}

\IEEEtitleabstractindextext{%
\begin{abstract}

Modeling heterogeneity by extraction and exploitation of high-order information from heterogeneous information networks (HINs) has been attracting immense research attention in recent times. Such heterogeneous network embedding (HNE) methods effectively harness the heterogeneity of small-scale HINs. However, in the real world, the size of HINs grow exponentially with the continuous introduction of new nodes and different types of links, making it a billion-scale network. Learning node embeddings on such HINs creates a performance bottleneck for existing HNE methods that are commonly centralized, i.e., complete data and the model are both on a single machine. To address large-scale HNE tasks with strong efficiency and effectiveness guarantee, we present \textit{Decentralized Embedding Framework for Heterogeneous Information Network} (DeHIN) in this paper. In DeHIN, we generate a distributed parallel pipeline that utilizes hypergraphs in order to infuse parallelization into the HNE task. DeHIN presents a context preserving partition mechanism that innovatively formulates a large HIN as a hypergraph, whose hyperedges connect semantically similar nodes. Our framework then adopts a decentralized strategy to efficiently partition HINs by adopting a tree-like pipeline. Then, each resulting subnetwork is assigned to a distributed worker, which employs the deep information maximization theorem to locally learn node embeddings from the partition it receives. We further devise a novel embedding alignment scheme to precisely project independently learned node embeddings from all subnetworks onto a common vector space, thus allowing for downstream tasks like link prediction and node classification. As shown from our experimental results, DeHIN significantly improves the efficiency and accuracy of existing HNE models as well as outperforms the large-scale graph embedding frameworks by efficiently scaling up to large-scale HINs.
\end{abstract}

}

\maketitle

\IEEEdisplaynontitleabstractindextext

%
\IEEEpeerreviewmaketitle

%

\IEEEraisesectionheading{\section{Introduction}\label{sec:introduction}}
Many real-world interactions recorded via digital medium take the form of heterogeneous information networks (HINs). HINs typically consist of multiple types of nodes connected by a variety of links/relationships. Due to the flexibility and diversity of relationships provided by HINs, heterogeneous network embedding (HNE) is widely adopted to perform several downstream tasks such as item recommendation, link prediction, and node classification. To effectively execute such tasks on an HIN, low-dimensional vector representations (i.e., embeddings) for each node are learned, such that the representations effectively preserve both feature and neighborhood information for each node belonging to that HIN. 

Advancing the accuracy of downstream tasks for HINs, a large number of centralized HNE approaches have been developed \cite{yang2020heterogeneous}, which learn the low-dimensional representation of the entire network on a single machine. However, in the case of large-scale HINs, these centralized frameworks at large neglect model scalability and efficiency.  

The centralized HNE paradigm requires the entire network data to be present and all model parameters to be stored and updated on one machine throughout the training phase. Though this assumption suits small networks, it makes centralized HNE models unable to scale to large networks due to multiple constraints from memory, computational cost, and accuracy perspectives. For instance, the graph transformer network (GTN) \cite{yun2019graph} needs to take the network's full adjacency matrix as its input, while a contemporary network can contain up to billions of nodes and edges (e.g. the WDC e-commerce dataset \cite{primpeli2019wdc}), making it unable to handle large-scale networks efficiently. Hence, attempts on improving the scalability of centralized HNE methods are made, such as the mini-batch implementation \cite{chen2018fastgcn} of graph convolutional network (GCN) \cite{kipf2016variational}, subgraph sampling \cite{hu2020heterogeneous}, and network coarsening \cite{liang2018mile}. However, the huge time consumption is unavoidable with sampled mini-batches/subgraphs as they need to be iterated over the whole network for model training \cite{chen2018fastgcn,hu2020heterogeneous}, and network coarsening \cite{liang2018mile} incurs accuracy loss as the condensed network fails to fully preserve the subtle characteristics of merged nodes/edges. Moreover, most of them \cite{chen2018fastgcn,liang2018mile} are designed purely for homogeneous networks, thus being inapplicable to HNE.

Another natural solution to embedding large-scale HINs is a decentralized HNE paradigm, that exploits \textit{model parallelism} \cite{shi2017frog} or \textit{data parallelism} \cite{imran2020decentralized}. In the case of model parallelism, the communication overhead renders it highly inefficient as billions of model parameters need to be shared and synchronized at each iteration to keep all learned node embeddings within the same distribution. In contrast, data parallelism is a more favorable choice for HNE as the network data can be partitioned and assigned to different workers, where each worker can deploy a fully localized model to learn node embeddings for a specific network partition.  In our previous work \cite{imran2020DDHH}, we presented a decentralized architecture DDHH based on data parallelism, that learns low-level vector representations of nodes on partitioned HINs and then align the independently learned representations onto the same space. Although DDHH efficiently computes network embeddings in a decentralized setting, it still suffers from two key challenges, which we will discuss below. 
\begin{flushleft}
\textbf{Challenge 1: Efficient Partitioning of HIN.} DDHH adopts a partitioning strategy where the two hyperedges (i.e., a set of nodes connected via a predefined relation) are contracted based on the number of edges and common nodes between them. This helps preserve the maximum node context within each partition. However, score comparison creates a bottleneck and prevents parallelism during the partitioning phase. Secondly, node links need to be recomputed after every iteration, increasing the search time significantly.
 
\textbf{Challenge 2: Context-aware Embedding Alignment.} Node embeddings learned through different workers need to be aligned onto a unified space to support subsequent tasks. For this purpose, DDHH uses an orthogonal alignment mechanism named Procrustes Alignment. DDHH, however, considers only a single learned representation of a node across multiple HIN partitions when learning the final embedding of each node, ignoring any other representations of that node present across multiple partitions. As each partition carries a specific context, the embedding alignment scheme sacrifices the expressiveness of learned node embeddings.
\end{flushleft}

In order to address the above challenges, we propose \textit{Decentralized Embedding Framework for Large-scale Heterogeneous Information Networks} (DeHIN) that significantly improves the DDHH model in our conference paper with fully distributed partitioning and context-aware alignment.
In order to compute partitions efficiently, we put forward a decentralized partitioning strategy based on hyperedges. The partitioning mechanism adopts a binary tree-like pipeline to match the nodes at different workers and a hash table that holds bucket ids for tracing each node's corresponding hyperedges, reducing the lookup time to a constant. 
In DeHIN, each worker has two buckets of nodes, and it contracts the hyperedges from one bucket into the other depending upon a similarity score. After each iteration, workers share the contracted bucket with their neighbors and the process is repeated until there is only one bucket containing all contracted partitions left. In this way, DeHIN effectively parallelizes the contraction-based HIN partitioning mechanism presented in DDHH while ensuring the quality of each partition. Meanwhile, instead of straightforwardly preserving a node's embedding from one partition during alignment, DeHIN adopts an aggregation strategy to preserve all contexts of each node by accounting for each of the network partitions it appears in. 
Specifically, as an extension to our conference paper \cite{imran2020DDHH}, this paper makes the following contributions:

\begin{itemize}
	\item We propose a novel, fully parallelizable HNE partitioning algorithm in DeHIN, which preserves the heterogeneity and high-order relationships in each generated partitions with the generated hyperedges while being highly efficient. 
	\item To maximize the contextual information within each learned node embedding, we design a context-aware alignment scheme. Based on translation matrix approximation, it maps the learned embeddings, received from the distributed workers, onto the a single orthogonal space by jointly considering a node's context in multiple partitions.
	\item To show the efficiency and effectiveness of DeHIN, we make use of 4 large-scale datasets, and conduct extensive experimentation on both effectiveness and efficiency. 
\end{itemize}


\section{Related Work}\label{sec:related}

In this section, we review related work regarding our research. Specifically, our work is relevant to heterogeneous network embedding, large-scale network embedding and hypergraph embedding.

\subsection{Network Embedding}
Network embedding assigns each node in the network a low-dimensional vector representation \cite{cui2018survey,zhang2018network}. These representations should effectively preserve the network structure and node properties \cite{imran2020decentralized}. Depending upon the type of nodes and links within a network, we can classify a network into two main categories \cite{zhang2019heterogeneous} i.e, homogeneous and heterogeneous networks, which require different embedding techniques. Another generic class of graph network representation is hypergraphs, which have recently been adopted to resolve a wide range of optimization problems, including scheduling, relation sampling among visuals, and video segmentation.
\subsubsection{Heterogeneous Network Embedding}
Heterogeneous networks are networks that contain different types of nodes and edges. Each node and edge belonging to a heterogeneous network tend to have unique attributes associated with it. These attributes are designed to capture different characteristics of that network \cite{shi2016survey,sun2012mining}. HNE models aim to sample a network's heterogeneity by either generating meta-paths such as metapath2vec \cite{dong2017metapath2vec}, HIN2Vec \cite{fu2017hin2vec} , MAGNN \cite{fu2020magnn},  HGT \cite{hu2020heterogeneous}  or by constructing smaller bipartite graphs such as PME \cite{chen2018pme}, HEER \cite{shi2018easing} and PTE \cite{tang2015pte}. They then apply transductive and inductive learning approaches \cite{cen2019representation,yang2020heterogeneous}, on the sampled data to obtain low dimensional vector representations for each node in the network. 

HNE models that adopt the bipartite strategy, initially split heterogeneous networks into multiple bipartite networks, by sampling both explicit and implicit links between the vertices \cite{gao2020learning,gao2018bine,huang2020biane}. They then apply different optimization strategies to learn node embeddings. A few HNE models that harness the structural utility of bipartite graphs also adopt centrality-based metric learning to capture high-order proximity \cite{chen2019exploiting,chen2018pme}.

\subsubsection{Hypergraph Embedding}

Hypergraph representation has been adopted as a tool to model complex high-order relationships, e.g. vertices in social networks possess far more complicated relationships than vertex-to vertex edge relationships among scheduling networks. Recently hypergraph embedding frameworks have been developed on top of homogeneous graph embedding models, including Spectral Hypergraph Embedding \cite{zhou2007learning}, HGNN \cite{feng2019hypergraph}, DHGNN \cite{jiang2019dynamic} and DHNE \cite{umeyama1988eigendecomposition}. Spectral Hypergraph Embedding \cite{zhou2007learning} treats network embedding as a k-way partitioning problem and uses eigendecomposition of adjacency matrix to approach the optimum matching problem. Similar to GCN, HGNN \cite{feng2019hypergraph} adopts spectral convolution to hypergraph and train the network using a semi-supervised learning approach. DHGNN \cite{jiang2019dynamic} stack multiple convolution layers, that use node and hyperedge features to compute the embeddings. After each layer, the output is used to reconstruct the hypergraph's structure, which is then passed on to the next layer. This is done to be better able to extract both global and local relationship information. DHNE \cite{umeyama1988eigendecomposition} uses an auto-encoder to preserve the structural information of hyperedges present in the network.

\subsection{Large-scale Network Embedding Frameworks}
%
%

\begin{figure}[t]
	
	\centering
	\includegraphics[width=\linewidth]{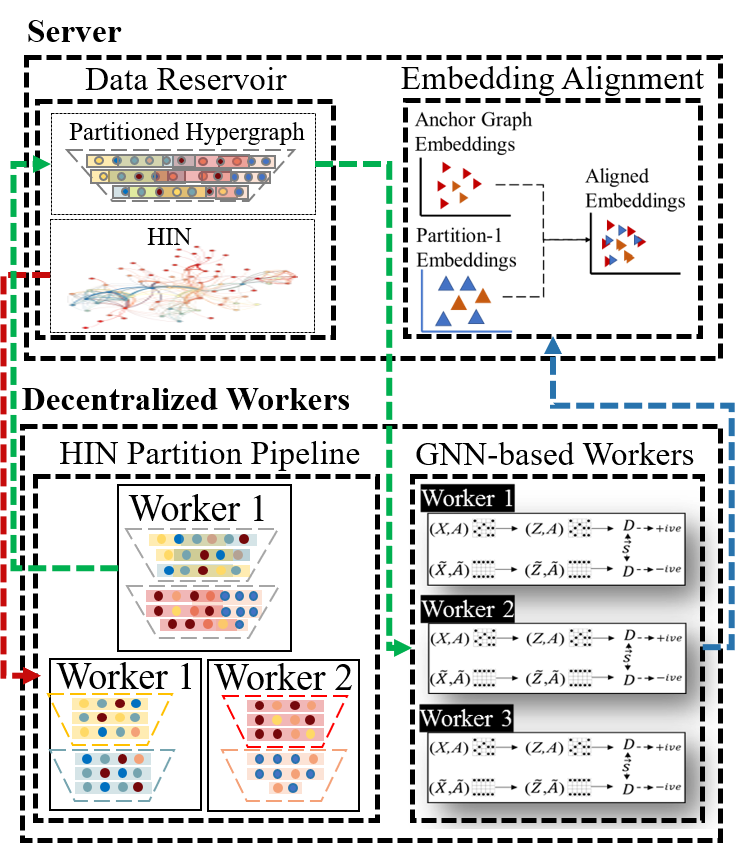}
	\caption{Pipeline of DeHIN. Multiple hyperedges are extracted from the HIN and placed in individual buckets. The workers iteratively contract two buckets to obtain the partitioned subnetworks. Each Partitioned subnetwork is assigned to an individual worker that learns embedding via mutual information maximization. The learned embeddings are then aligned at the server side through context-aware orthogonal procrustes transformation.
	}
	\label{framework}
\end{figure}

Most existing node embedding models are designed to process networks having a few thousand nodes \cite{zhu2019graphvite}. However, real-world networks are usually very large. Scaling the existing network embedding models to efficiently process networks with tens of millions of nodes is a challenging problem. In order to meet this demand, recently a few large-scale network embedding frameworks have been purposed, which can adapt to very large networks on a single machine.

Among the large-scale network embedding frameworks, MILE \cite{liang2018mile}, iteratively coarsens the network to execute network embedding on a much smaller, contracted network. The embeddings of the contracted nodes are refined to the original nodes using a deep model. A key limitation of MILE is its ability to adapt with scaling network size. In this approach, the embedding quality is greatly affected, by the increase in network size. Additionally, the embedding refinement mechanism of the source node, down from a super node is a computationally expensive process for large-scale networks. Parameter sharing, i.e., model parallelism, is another approach to deal with very large-scale datasets. COSINE and Py-Torch-BigGraph (PGB) \cite{pbg,zhang2018cosine} utilizes parameter sharing approach. These models generate non-overlapping partitions, having distinct vertices.  The limitation posed to these large-scale network embedding frameworks is that they are highly dependent on the bus bandwidth. 

Another very widely adopted large-scale network embedding framework is GraphVite \cite{zhu2019graphvite}. It is a hybrid system that utilizes both, the CPU and the GPU for training node embeddings. Training is made by utilizing and co-optimizing the existing network embedding algorithm. They also purpose an efficient collaboration strategy is to further reduce the cost of synchronization among CPUs and GPUs. DeLNE \cite{imran2020decentralized} adopts data parallelism approach to deal with large-scale networks. It partitions a network into smaller part \cite{karypis1999multilevel} and learn vector representation in a completely decentralized fashion, using Variational Graph Convolution Auto-Encoders (VGAE) \cite{kipf2016variational}. 

Although the above-mentioned large-scale network embedding frameworks are efficient and produce quality embeddings as reported in \cite{imran2020decentralized}, these frameworks are designed for homogeneous networks. Heterogeneous network embedding models require high-order characteristics of a node to be preserved along with its structure, during the structure sampling phase. Secondly, partitioning a heterogeneous network, that consists of interconnected nodes and edges of different types, is a far more complex and tedious task. PBG \cite{pbg} is a distributed system that trains on an input network by ingesting its list of edges and their relation types. However, the PBG still requires a global model to be every iteration. This constant sharing of information, as depicted in our results, is extremely inefficient as the number of links grows. Decentralized Deep Heterogeneous Hypergraph (DDHH) \cite{imran2020DDHH} is another decentralized node embedding framework, specifically designed for HNE, that uses hyperedge sampling to partition HIN. Each worker in DDHH uses the information maximization theorem for learning node embeddings locally then aligns the learned embeddings using an orthogonal alignment mechanism. However, as mentioned in Section \ref{sec:introduction}, it suffers from two major drawbacks i.e., it incurs high computational and time cost in its partitioning mechanism, and also loss of high order context during alignment. 
To address the efficiency issue of DDHH, we propose a fully parallelizable HNE partitioning algorithm in DeHIN. DeHIN also introduces a  hash table that keeps track of each node's hyperedges, efficiently providing the best contraction option and discards the need for recomputing/relocating node's links at each iteration. To preserver high-order context during the alignment phase, DeHIN maximizes the contextual information within each learned embeddings, using a context-aware alignment scheme.


\section{Preliminaries}
\label{sec:preliminaries}
Here, we introduce some of the preliminary concepts that are commonly adopted throughout the paper and formulate the research problem.

\begin{flushleft}\textbf{Definition 2.1:} A \textbf{HIN} consist of vertices belonging to one or more types, connected via one or more types of edges. Formally, an HIN can be represented as $G = \{\mathcal{V,E,X,R}\}$. Here $\mathcal{V}$ is the set of all vertices $v$  type and $\mathcal{E}$ is the set of all edges, connecting nodes $e = (v, v', r)$, through relationship type $r\in\mathcal{R}$, and $\mathcal{X}$ represents the set of node features $\textbf{x}$. 
\end{flushleft}

\begin{flushleft}\textbf{Definition 2.2:} A \textbf{connected component} (CC) represents a set of nodes $\{v_1,v_2,v_3...,v_n\}$ connected via single path $p_i$, such that all the edges $\{e_1,e_2,e_3...,e_n\}$  in $p_i$ belong to the same relationship type $r$.\end{flushleft}

\begin{flushleft}\textbf{Definition 2.3:} A \textbf{hyperedge} $e^H = \{v_1,v_2,...,v_n\}$ can connect two or more vertices simultaneously. Given an HIN, we contract all connected vertices $\mathcal{V}$ having same relationship types $r\in \mathcal{R}$, into multiple sets. Each set represent hyperedges $\mathcal{E}^H$, where each $v$ in a hyperedge $e^H$ have common properties. A $v$ is incident to a hyperedge $e^H$  if $v \in e^H$.
\end{flushleft}

\begin{flushleft}\textbf{Definition 2.4:} A \textbf{hypergraph} is a graph $\mathcal{G}^H = \{\mathcal{V},\mathcal{E}^H\}$ having vertices $\mathcal{V}$ connected via hyperedges $\mathcal{E}^H$. An HIN is converted into a hypergraph by generating all possible  hyperedges $\mathcal{E}^H$, as described in definition 2.3.
\end{flushleft}

\begin{flushleft}\textbf{Definition 2.5:} An \textbf{anchor network} $\mathcal{G}^A = \{\mathcal{V}^{A},\mathcal{E}^{A}\}$ contains such vertices $v$ that are shared among two or more subnetwork. These vertices are called  anchor (i.e., common) nodes $\mathcal{V}^{A}\in \mathcal{V}$. First-order neighbours of each anchor node is also included in the anchor network $\mathcal{V}^A$, connected via hyperedges $\mathcal{E}^{A}$.

\end{flushleft}
\begin{flushleft}\textbf{Problem Definition:} \textbf{Decentralized Embedding for Large-scale HIN}. Given an HIN, our goal is to compute a low-dimensional vector representation $\textbf{v} \in \mathbb{R^\textit{d}}$ for each vertex $v\in \mathcal{V}$ in $\mathcal{G}$, by aborting distributive partitioning approach, that partitions $\mathcal{G}$ in into $k$ subnetworks $\{\mathcal{G}_1,\mathcal{G}_2,...,\mathcal{G}_k\}$ and learn node embeddings for each subnetwork independently. Finally, the independently learned embeddings of all partitions are eventually mapped onto the same vector space via an alignment scheme $\mathcal{M}_{\mathcal{G}_n}(\widehat{\textbf{Z}})$.\end{flushleft}

\section{Methodology}
\label{sec:methodology}
The decentralized learning pipeline of DeHIN in Fig. \ref{framework}  can be divided into three phases. First of all, a partitioning unit generates hypergraphs from the given HIN and then adopts a parallel partitioning approach to obtain multiple smaller subnetworks. Secondly, at each worker, we utilize a deep information maximization algorithm to learn vector representations of largely unlabeled nodes. And Finally, a server-side that aligns all the embeddings received from each worker, into a unified space. Details of each component are introduced in the following section.


\subsection{HIN Partition}
In order to effectively preserve the structural and higher-order properties of a node, GNNs require full access to the node's neighborhood belonging to each of its relationships $\mathcal{R}$. However real-world HINs are billion-scale in size, containing millions of nodes and edges of numerous types. Learning quality node representations for such HINs, by adopting a framework based on centralized or model parallelism architecture, is extremely infeasible as explained in Section \ref{sec:introduction}. Therefore, we devise a smart data parallelism framework that allows the HIN to be efficiently partitioned and re-aligns the independently learned node embeddings.

To better retain the network structure and high-order information for the node embedding task, each partition should: 1) \textit{maximize the diversity of each node's relationship}, so that each partition retains (almost) all possible behaviors of its nodes, 2) \textit{maintain each node's neighborhood}, so that the network structure is preserved. Our proposed HIN partition pipeline consists of two main steps, namely Hypergraph Generation and Distributed Network Partition. 

\subsubsection{Hypergraph Generation}\label{hyp_gen_subsection}
Heterogeneous network exhibits  semantically complex and high-order relationships among its nodes, e.g., in IMDB network, a \textit{viewer} can review a \textit{movie}, and this \textit{movie} is directed by a \textit{director} \cite{maas-EtAl:2011:ACL-HLT2011}. For most GNN-based HNE approaches, a heterogeneous network is commonly input into the models in three ways: 1) in the form of an adjacency matrix of the whole network \cite{yun2019graph}, 2) by converting it into relation-specific bipartite graphs \cite{wang2019neural} or 3) by creating problem-specific metapaths \cite{shi2018heterogeneous}. However, when the size and diversity grow on larger heterogeneous networks, the first two methods incur heavy memory footprints, thus being unsuitable for embedding large networks. Though metapath-based sampling can be modified for generating network partitions, it has limited flexibility 

Intuitively, in an HIN, different relationships $r\in \mathcal{R}$ among nodes represent a certain perspective \cite{wang2019heterogeneous}, i.e., a node's role in context to its varying relationship with its neighbors. For example, in a publication dataset \cite{tang2008arnetminer}, a paper serves as a citation in a \textit{paper-paper} relationship, while it is a publication record in a \textit{paper-author} relationship. Hence, based upon the relationship a node has with its neighbors, we identify and construct hyperedges. Specifically, we adopt an edge contraction strategy to sample meaningful hyperedges from an HIN. All the edges belonging to a $r\in \mathcal{R}$ are contracted, to form hyperedges of type $r$, and this process is carried out in parallel for each relationship type in $\mathcal{R}$.

\begin{figure}[t]
	\centering
	\includegraphics[width=6.2cm]{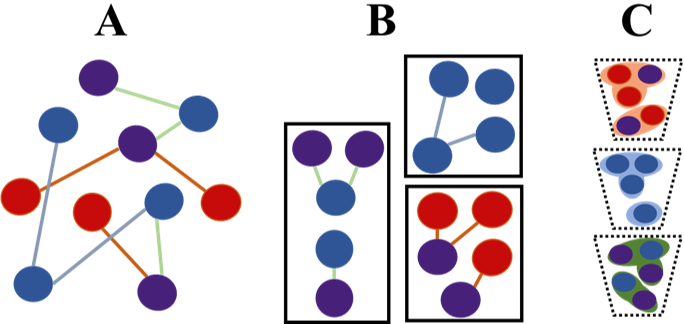}
	
	\caption{ \textbf{A:} The original HIN.\textbf{ B:} Each edge type is sampled into different buckets. \textbf{C:} All connected nodes in each bucket form hyperedges.
	}
	\label{hyperedge_generation}
	\vspace{-4.5mm}
\end{figure}

In DeHIN, sampling hyperedges by contraction is motivated by the graph's connectivity, as connected nodes (e.g., nodes in a pairwise relationship or a path) usually possess high proximity compared with disconnected nodes \cite{gilpin2013guided,imran2018exploiting}. In other words, hyperedges are sets of connected vertices extracted from each relationship type within a heterogeneous network. Given an HIN  $\mathcal{G}$ having $|\mathcal{V}|$ nodes, $|\mathcal{E}|$ edges and $|\mathcal{R}|$ relationships, we generate $|\mathcal{R}|$ buckets $\mathcal{B}$, each holding a collection of hyperedge belonging to a single relation $r \in \mathcal{R}$.
Essentially, we engage $|\mathcal{R}|$ workers so that all nodes connected via same $r \in \mathcal{R}$ (i.e. all nodes in $\mathcal{G}^r$) are split into separate buckets, as shown in Fig. \ref{hyperedge_generation}\textbf{B}. Nodes in each bucket are then contracted by utilizing parallel Breadth First Search (BFS). These collected paths, representing connected components in $\mathcal{G}^r$, are denoted as hyperedges $\mathcal{E}^H \in \mathcal{B}^r$ which is illustrated in Fig. \ref{hyperedge_generation}\textbf{C}. 

To this end, $\mathcal{E}^H$ encloses complete high-order neighbourhood set of each contained node, within a single $r$.  The time complexity of generating hyperedges for $|R|$ buckets in parallel is  $\mathcal{O}(|\mathcal{E}^r_{max}|) $, where $|\mathcal{E}^r_{max}|$ represents the highest number of edges in any $r$. The space complexity of this module is $\mathcal{O}(|\mathcal{E}^r|+|\mathcal{V}^r|)$, where $\mathcal{E}^r$ and $\mathcal{V}^r$ represents the total number of edges and vertices on the worker with relationship $r$.
%

\subsubsection{Distributed Network Partition}
\begin{figure}[!t]
	\centering
	\includegraphics[width=8cm]{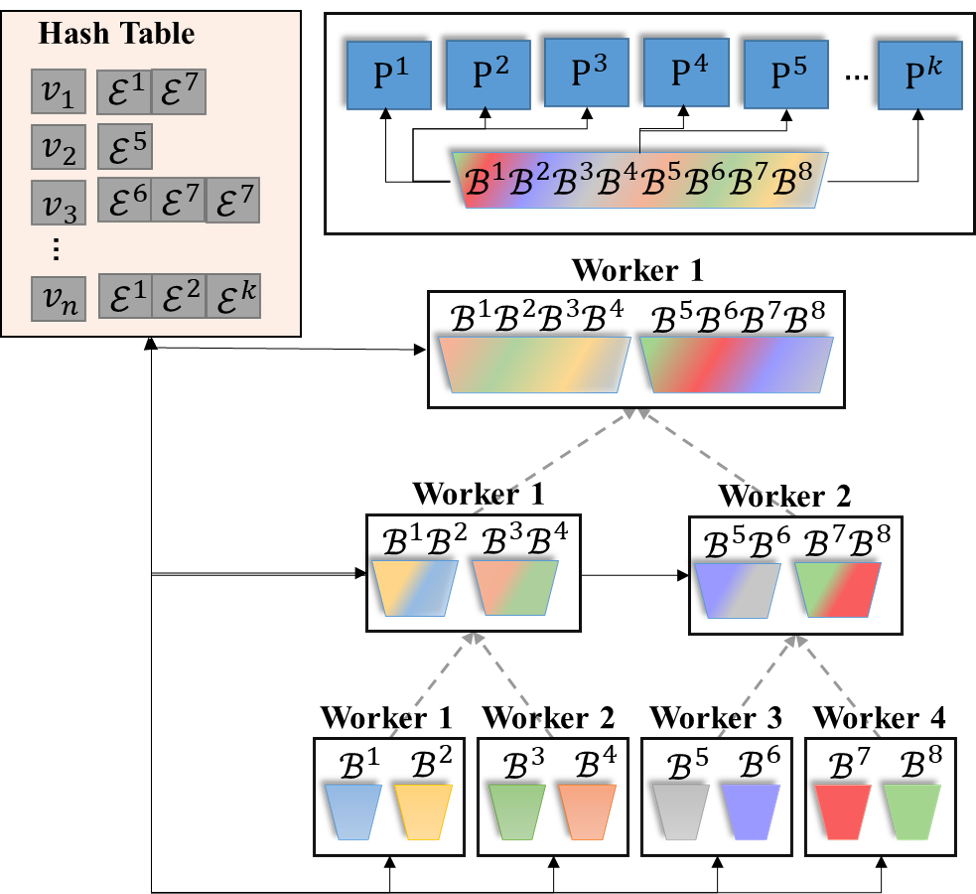}
	\caption{Iterative contraction of $\mathcal{E}^H$ between buckets. Hash table maintains all the hyperedges a node is engaged in. }
	\label{buckets}
	\vspace{-2mm}
\end{figure}

The main objective of our partition task is to produce semantically distinct subnetworks and ensure parallelism among the distributed workers. The hyperedge construction strategy we adopted in Section \ref{hyp_gen_subsection} ensures that each hyperedge encompasses the complete single-typed neighborhood of each node. This property helps to preserve the structure of the HIN. We design a scoring function in Algorithm 1 that makes sure a maximum number of $|\mathcal{R}|$ of a node are grouped together. We also provide a graphical explanation in Fig. \ref{buckets}. Initially $\frac{|\mathcal{B}|}{2}$ workers are generated. Each worker is assigned two buckets each, denoted by $\mathcal{B}_1$ and $\mathcal{B}_2$. Hyperedges ($\mathcal{E}^H \in \mathcal{B}_1$) are sorted on the basis of their size. We use Red and Black (RB) trees as a data structure of the buckets for efficiency. A scoring function, that uses a hash table to find which $\mathcal{E}^H \in \mathcal{B}_2$ that has most common nodes to $\mathcal{E}^H \in \mathcal{B}_1$. Those $\mathcal{E}^H \in \mathcal{B}_2$ having more common nodes with $\mathcal{E}^H \in \mathcal{B}_1$, are contracted from smallest to largest. The hash table points to the list of hyperedges ($\mathcal{E}^H$) each node is affiliated with. This scoring function makes sure that maximum $\mathcal{R}$ are grouped together. In the next iteration $\frac{|\mathcal{B}|}{2}$ (with updated $|\mathcal{B}|$) workers are generated, and the same process is repeated until only one bucket remains. 

\begin{figure}[t]
	\centering
	\includegraphics[width=\linewidth]{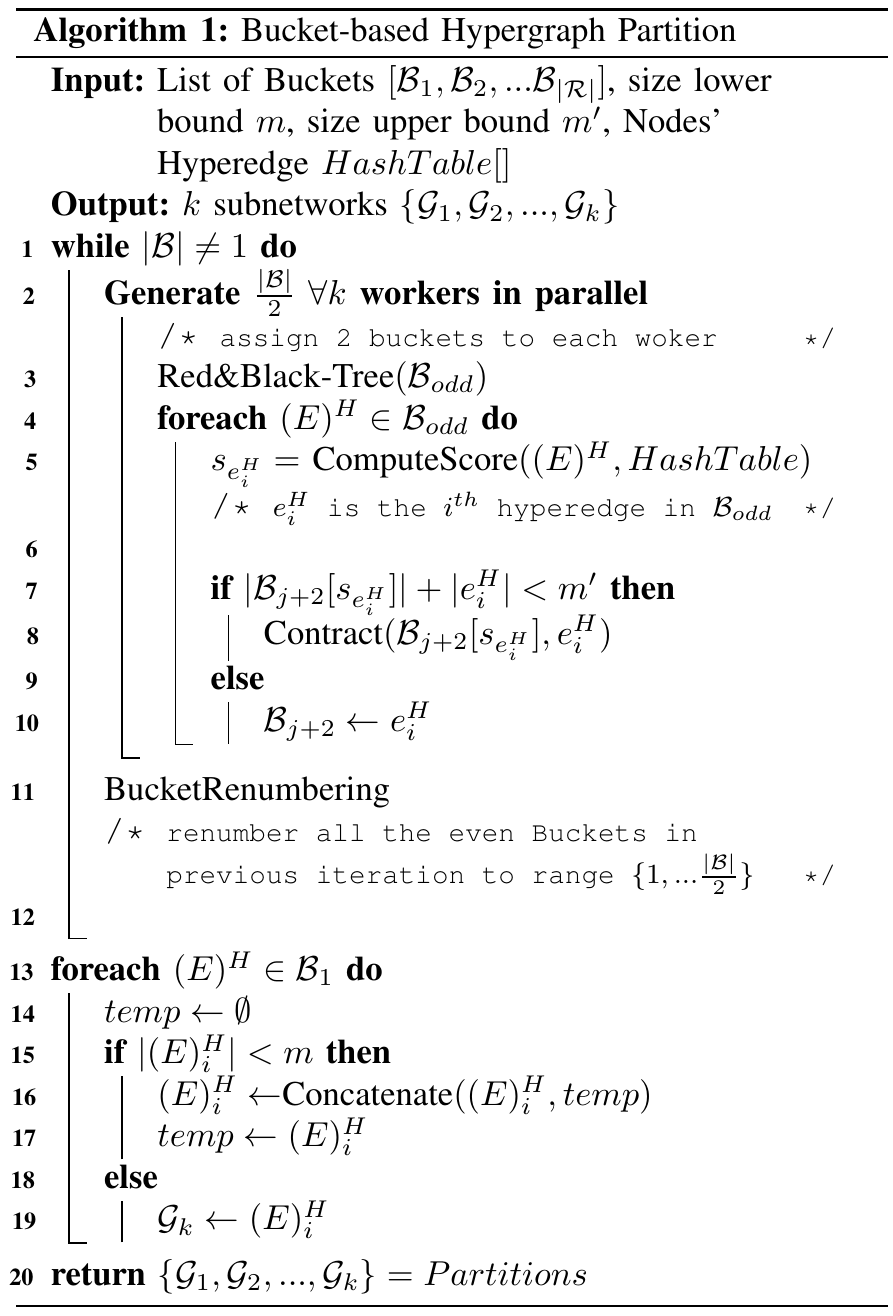}
	\label{algo_partition}
	\vspace{-0.5 cm}
\end{figure}

In order to avoid partition size implosion or explosion, we always set a lower-bound and an upper-bound for the contractions. Two hyperedges are contracted only if the upper-bound is not exceeded. In case a single hyperedge exceeds the upper bound, it no longer takes part in the contraction process and remains deactivated for remaining iterations. When only one bucket is left, all the hyperedges that are smaller than the lower-bound are contracted together randomly. For our experiments, we set the lower-bound to 10,000 nodes. Since each partition encloses the complete neighborhood of a node w.r.t. at least one $r$, our opted lower-bound is sufficient for the model to preserve neighborhood structure. High-order information of these nodes is comprehended by the alignment mechanism, given in Section \ref{sec:EA}. To obtain a reasonable data size for our distributed embedding models, we keep the upper-bound of the contracted $\mathcal{E}^H$ to a maximum value between the lower-bound and the largest connected component  (the size of largest $\mathcal{E}^H$). Worst-case running time of Algorithm 2 is $\mathcal{O}(k|\mathcal{B}_k||\mathcal{E}^H|\log|\mathcal{E}^H|)$, where $k$ is the number of iterations and $\mathcal{B}_k$ is the smallest bucket in the $k^{th}$ iteration. The space complexity at a given worker $b$ during this stage is $\mathcal{O}(|\mathcal{V}_b\mathcal{E}^H_b|)$, where $\mathcal{V}_b$ represents the vertices on worker $b$ and $\mathcal{E}^H_b$ are the total number of hyperedges connecting those vertices.

Finding buckets to be contracted at each iteration, such that for every node we can sample most of its relations, without violating upper-bound, is an NP-Hard problem. Since GNNs can easily learn structural properties of a node in single $r \in \mathcal{R}$ and our alignment scheme can effectively align heterogeneous node contexts from different partitions, our context-aware neighborhood partition is sufficient.

\newcommand{\R}{\mathbb{R}}

\subsection{Distributed Subnetwork Embedding}\label{sec:DSE}
Each subnetwork constructed thus far contains a number of connected components and is rich with neighborhood semantics. These subnetworks are then assigned to a distributed worker, which computes node embeddings using a local model. The main objective of each worker model is to maximize information between the connected nodes of the provided subnetwork. We denote union of each node's $v_i \in \mathcal{G}_n$ ($1\leq n \leq k$) first-order neighborhood as $\mathcal{S}_i$ and term its corresponding hyperedge as a patch $\mathcal{P}_i = \mathcal{S}_v\cup \mathcal{E}^{H}_i$.

GNNs have been widely adopted as an effective approach for embedding heterogeneous networks \cite{yun2019graph,wang2019heterogeneous}. Utilizing the power of GNNs, in our work, we approach the node embedding task as an unsupervised representation learning problem without any assumptions on the availability of labeled nodes. 
To achieve this, any GNN model can be adopted at this stage. DeHIN provides the flexibility to choose any network embedding model during subnetwork embedding stage. This includes models with supervised loss functions such as hinge loss or link prediction-based loss in the absence of labeled data. However, in our work, we deal with large-scale networks having a very small amount of labeled data. Secondly, we deal with heterogeneous data where the links between nodes are of multi-type.
However, in our work we exploit the theorem that two nodes should exhibit high affinity if they are similar in their features or structural properties. By taking advantage of similarities between the neighborhood structure and the node features in a subnetwork, we make use of deep graph information maximization (DGIM) \cite{velickovic2019deep} based approach for unsupervised learning task, where we substitute hyperedges  as patches. To learn a node's embedding representation, the DGIM model maximizes the mutual information between a node's patch $\mathcal{P}_i$ and the summary of the complete subnetwork $\mathcal{G}_n$. Here $\mathcal{P}_i$ represents a hyperedge and $\mathcal{G}_n$ represents the collection of all hyperedges belonging to a single subnetwork. 

We construct a feature matrix $\textbf{X}\in \mathbb{R}^{N\times d_0}$ for a $\mathcal{G}_n$, having $N=|\mathcal{G}_n|$ nodes. The feature vector of  $v_i\in\mathcal{G}_n$ is represented by the $i$-th row $\textbf{x}_i\in \mathcal{X}$. To mathematically represent the connections among the nodes in $\mathcal{G}_n$, we also define an adjacency matrix  $\textbf{A}\in \mathbb{R}^{N\times N}$, having self-loops.
Our main objective is to learn an encoder function $\mathcal{F}_{\mathcal{G}_n}(\cdot)$, such that $\mathcal{F}_{\mathcal{G}_n}(\textbf{X},\textbf{A}) = \textbf{Z} \in \mathbf{R}^{N\times d} = [\textbf{z}_1;\textbf{z}_2;...;\textbf{z}_n]$ where matrix $\textbf{Z}$ carries all $N$ nodes' embedding:
\begin{equation}
\textbf{Z}=\mathcal{F}_{\mathcal{G}_n}(\textbf{X},\textbf{A}).
\label{equ:gm1}
\end{equation}
In DeHIN, we design the encoder using graph convolution network (GCN) \cite{kipf2016variational} for effective  aggregation of the node's neighborhood. GCN assists the learning process by effectively aggregating a node's neighborhood embeddings, accumulating features and structural information within the embedding. The propagation rule for a $l$ layered GCN, where $l\geq 1$, is as follows:
\begin{equation}
\textbf{Z}^{(l)} = \mathcal{F}_{\mathcal{G}_n}(\textbf{Z}^{(l-1)},\textbf{A})=\sigma(\textbf{D}^{-\frac{1}{2}}\textbf{A} \textbf{D}^{-\frac{1}{2}}\textbf{Z}^{(l-1)}\textbf{W}^{(l-1)}),
\label{equ:gcn}
\end{equation}
where $\sigma(\cdot)$ represents a nonlinear activation function and $\textbf{D}$ is a diagonal degree matrix that is  normalized by $-\frac{1}{2}$ power. The embedding and the weight matrices of all nodes in $\mathcal{G}_n$, for layer $l$, are represented as $\textbf{Z}^{(l)}$ and $\textbf{W}^{(l)}$. Here, for layer 0 we set $\textbf{Z}^{(0)}=\textbf{X}_i$. We denote the $i$-th node embedding in the final output $\textbf{Z}^{(l)}$ as $\textbf{z}_i$. Essentially, each node representation $\textbf{z}_i$ (i.e., the $i$-th row of $\textbf{Z}$) can be viewed as a patch-level summary for $\mathcal{P}_i$. 

In our information maximization scheme, we aim to maximize the probability of having the learned patch-level summary $\textbf{z}_i$ belong to the subnetwork $\mathcal{G}_n$. In this regard, we obtain the subnetwork-level global summary vector $\textbf{s} \in \mathbb{R}^{1\times d}$ with a readout function $\mathcal{R}(\cdot)$, which combines all $\textbf{z}_i\in \textbf{Z}^{(l)}$ into a unified representation. As the learned $\textbf{z}_i$ is sufficiently expressive, we formulate $\mathcal{R}(\cdot)$ as the following average pooling with nonlinearity to avoid excessive computational complexity:
\begin{equation}
\textbf{s} = \mathcal{R}\left(\mathcal{F}_{\mathcal{G}_n}(\textbf{X},\textbf{A})\right) = \sigma(\frac{1}{N}\sum_{i=1}^{N}\textbf{z}_i).
\label{equ:summary_vector}
\end{equation}
With the subnetwork representation $\textbf{s}$, we utilize a discriminator function $\mathcal{D}(\cdot)$, such that $\mathcal{D}(\textbf{z}_i,\textbf{s})$ outputs a scalar indicating the affinity score between each $(\textbf{z}_i,\textbf{s})$ pair. In order to maximize $\mathcal{D}(\textbf{z}_i,\textbf{s})$, we also generate corrupted patch representation w.r.t. each node $v_i$, denoted as $\widetilde{\textbf{z}}_i$. Therefore, a fully-trained $\mathcal{D}(\cdot)$ can be expected to produce a higher pairwise score for a positive tuple $(\textbf{z}_i,\textbf{s})$, and a lower one for the negative tuple $(\widetilde{\textbf{z}}_i,\textbf{s})$. Specifically, by corrupting either the original node connections or raw node features in $\mathcal{G}_n$, we can naturally produce the corrupted representation $\widetilde{\textbf{z}}_i$ with Eq. \ref{equ:gcn} as each generated $\widetilde{\textbf{z}}_i$ no longer reflects the authentic information about network structure or node properties. In DeHIN, we generate corrupted edges, represented by $\widetilde{\textbf{A}}$, by  inserting new edges and deleting existing edges from $\widetilde{\textbf{A}}$ at random. Note that though each node's feature vector remains unchanged, we denote the corresponding node feature matrix as $\widetilde{\textbf{X}}\in \mathbb{R}^{\widetilde{N}\times d_0}$ because some nodes might vanish with deleted edges, returning a total of $\widetilde{N} \leq N$ nodes.

As an objective function, we adopt binary cross-entropy loss  for all positive and negative instances:
\begin{equation}
\begin{split}
\mathcal{L} & = \frac{1}{N+\widetilde{N}}  
\Big{(}\sum_{i=1}^{N} \mathbb{E}_{(\textbf{X},\textbf{A})} \left[\log\mathcal{D}(\textbf{z}_i,\textbf{s})\right] \\
&+ \sum_{j=1}^{\widetilde{N}} \mathbb{E}_{(\widetilde{\textbf{X}},\widetilde{\textbf{A}})} \left[\log\mathcal{D}(\widetilde{\textbf{z}}_j,\textbf{s})\right]\Big{)}\label{eq1:loss},
\end{split} 
\end{equation}
where we formulate the discriminator $\mathcal{D}(\cdot)$ as a feedforward network followed by a sigmoid-activated projection layer:
\begin{equation}
\mathcal{D}(\textbf{z}_i,\textbf{s}) = \textnormal{sigmoid}\big{(}\sigma([\textbf{z}_i,\textbf{s}]\textbf{W}_\mathcal{D} + \textbf{b})\cdot \textbf{w}^{\top}\big{)},
\end{equation}
where $\textbf{W}_{\mathcal{D}} \in \mathbb{R}^{2d\times d}$ and $\textbf{b}\in \mathbb{R}^{1\times d}$ are the weight and bias, while $\textbf{w}\in \mathbb{R}^{1\times d}$ is the projection weight.
In Eq. \ref{eq1:loss}, based on Jensen-Shannon divergence \cite{lin1991divergence} between the joint (i.e., positive instances) and the product of marginals (i.e., negative instances), thus thoroughly maximizing the mutual information between $\textbf{z}_i$ and $\textbf{s}$. The subnetwork embedding model is trained via stochastic gradient descent to minimize loss $\mathcal{L}$. As such, we can obtain high-quality node embeddings for partition $\mathcal{G}_n$. For all $k$ subnetworks $\{\mathcal{G}_1, \mathcal{G}_2,...,\mathcal{G}_k\}$, each worker has the same learning scheme but the parameters are specific to each partition $\mathcal{G}_n$ and are not shared across different partitions. So, the distributed embedding model on each worker is fully localized, and no communication costs are introduced in the subnetwork embedding phase.

\subsection{Embedding Alignment}\label{sec:EA}
As we create an independent embedding model for each partition, the corresponding node representations are independently learned from a partition-specific vector space. Also, different partitions may overlap in some common nodes in different contexts i.e. relationship type $\mathcal{R}$, leading to multiple embeddings in different context for the same node. To perform subsequent tasks, all node embeddings from different subnetworks need to be projected onto the same vector space, so as to be positioned in the same distribution and become comparable to each other. One naive approach is constructing pairwise mappings between the embeddings of adjacent partitions or by translating the embeddings of all $k$ partitions onto the same space.  
However, transferring and mapping all the nodes onto the same space is a computationally exhaustive process. Another embedding alignment approach, commonly adopted in cross-lingual word embedding tasks, is projecting the embeddings of source subnetwork onto target subnetwork using anchor (common) nodes. The problem with this approach is that it does not account for heterogeneous data and the resulting embeddings are not on the same orthogonal space. 
Hence, we propose an efficient yet effective embedding alignment approach for those distributively learned node embeddings. 

Rather then transferring and mapping all the node onto the same space, which is a computationally exhaustive process, we present an efficient mapping scheme.
During the partition phase we sample the top $k$ number of nodes from each subnetwork which are connected to one or more subnetworks, along with their first-order neighborhood. Here top nodes refers to the nodes having maximum out degree with the neighboring subnetworks, and $k$ is determined by the lower and upper-bound of the newly formed subnetwork. We refer to this subnetwork as anchor network and is defined as $\mathcal{G}^A=\{\mathcal{V}^A, \mathcal{E}^A\}$. We learn node embeddings of $\mathcal{G}^A$ locally on the server using a DGIM model and the corresponding vector space of anchor network embeddings $\textbf{Z}^A$ is regarded as the target space. Our aim is to align all the incoming subnetwork embeddings onto anchor network's embedding space, by learning mapping functions $\mathcal{M}_{\mathcal{G}_n}(\cdot)$.

Consider a subnetwork $\mathcal{G}_n$ having $\mu$ anchor nodes from $\mathcal{G}^A$. Aim of the learned mapping function $\mathcal{M}_{\mathcal{G}_n}(\cdot)$ is to: 1) project  $\mu$ anchor node embeddings onto the target vector space; and 2) ensures the projected embeddings preserve high similarity with the same anchor node embeddings learned from $\mathcal{G}^A$. Using the learned $\mathcal{M}_{\mathcal{G}_n}(\cdot)$, we are able to map all the node embeddings from $\mathcal{G}_n$ onto the same space as that of $\textbf{Z}^A$, which is the target vector space. This not only enables us to perform embedding alignment on each individual subnetwork embedding with substantially lowered computational power and storage requirements, but also makes the embedding process highly salable.


\begin{figure}[t]
	\centering
	\fbox{\includegraphics[width=8.6cm]{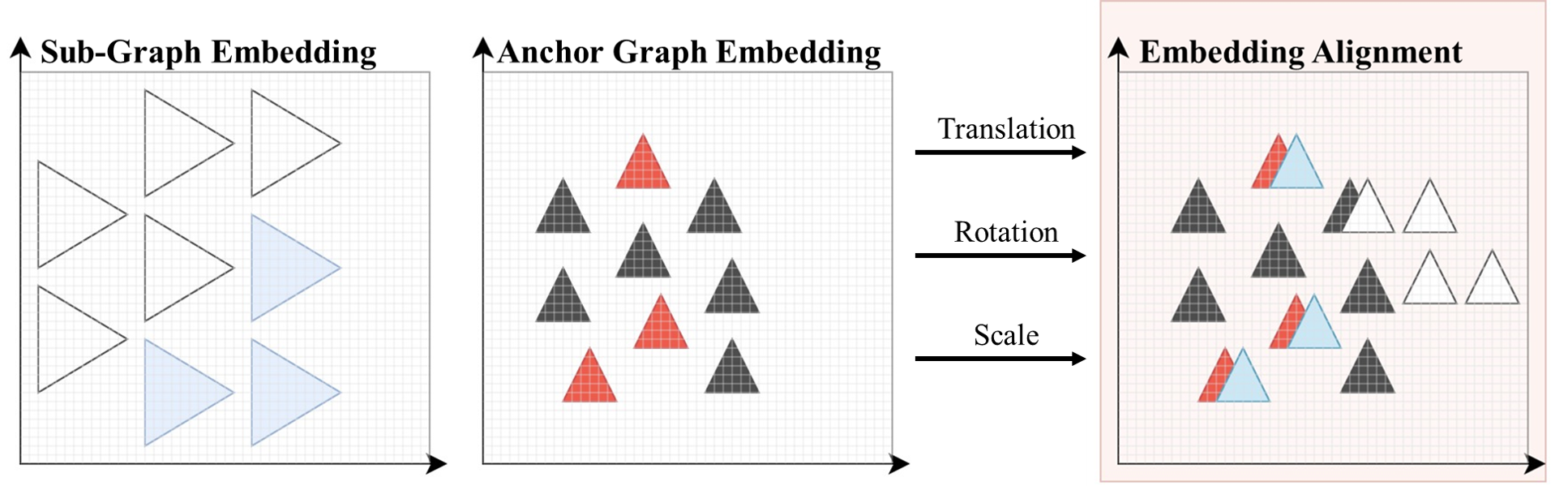}}
	\vspace{-5mm}
	\caption{The embedding alignment process. There are three common nodes shared between the subnetwork (blue triangles) and the anchor graph (red triangles). The embeddings of these nodes are aligned by a mapping function consisting of rotation, scaling, and translation. The remaining nodes belonging to the source subnetwork are mapped onto same orthogonal space with the learned mapping function.
	}
	\label{fig:alignment}
	\vspace{-2 mm}
\end{figure}

We further elaborate our alignment mechanism by considering a single partition $\mathcal{G}_n$. We collect all $\mu$  anchor node embeddings $\{\textbf{z}_i|\forall v_i \in \mathcal{G}_n \cap \mathcal{G}_A\}$ from $\mathcal{G}_n$.

From its node embeddings $\{\textbf{z}_1, \textbf{z}_2,...,\textbf{z}_N\}$, we gather all $\mu$ anchor node embeddings $\{\textbf{z}_i|\forall v_i \in \mathcal{G}_n \cap \mathcal{G}_A\}$.
Using the approach discussed in Section \ref{sec:DSE}, we can obtain node embeddings $\{\textbf{g}_1, \textbf{g}_2, ..., \textbf{g}_{|\mathcal{V}^A|}\}$ (of the same embedding dimension $d$) for the anchor network $\mathcal{G}^A$, and acquire set $\{\textbf{g}_i|\forall v_i \in \mathcal{G}_n \cap \mathcal{G}_A\}$. To support subsequent computation, we convert both sets of anchor node embeddings into matrices, respectively denoted by $\widehat{\textbf{Z}}, \widehat{\textbf{G}} \in \mathbb{R}^{\mu \times d}$. Then, we formulate $\mathcal{M}_{\mathcal{G}_n}(\cdot)$ as a sequence of transformations on $\widehat{\textbf{Z}}$ such that $\mathcal{M}_{\mathcal{G}_n}(\widehat{\textbf{Z}})\approx \widehat{\textbf{G}}$. To be specific, as shown in Fig. \ref{fig:alignment}, the transformations include rotation and scaling on all anchor node embeddings in $\widehat{\textbf{Z}}$, followed by a vector-level translation:
\begin{equation}
\label{eqn:trans}
\mathcal{M}_{\mathcal{G}_n}(\widehat{\textbf{Z}}) = c\cdot \widehat{\textbf{Z}}\textbf{T}+\textbf{j}^{\top}\textbf{t} \approx \widehat{\textbf{G}},
\end{equation}
where $\textbf{T} \in \mathbb{R}^{d \times d}$ is the rotation matrix, $c \in \mathbb{R}$ is the scaling factor, $\textbf{t} \in \mathbb{R}^{1\times d}$ is the translation vector and $\textbf{j} = \textbf{1}^{1\times \mu}$ is a vector consisting of $1$s. To ensure $\widehat{\textbf{Z}}$ is mapped onto the same orthogonal space as $\widehat{\textbf{G}}$, we quantify the alignment loss as follows:
\begin{equation}
\label{eqn:op1}
\Psi = \left\|c \cdot \widehat{\textbf{Z}}\textbf{T}-\widehat{\textbf{G}}\right\|_F+\textbf{j}^{\top}\textbf{t},
\end{equation}
where $\left\|\cdot\right\|_F$ is the Frobenius norm. In order to estimate those three parameters in $\mathcal{M}_{\mathcal{G}_n}(\cdot)$, we employ least square correspondence \cite{gower1975generalized,ten1977orthogonal} in DeHIN. To obtain the least square estimate of $\textbf{T},\textbf{t}$ and $c$, we devise the following Lagrangean expression \cite{son2012some}:
\begin{equation}
\textbf{F} = tr(\Psi^{\top}\Psi)+tr(\textbf{L}(\textbf{T}^{\top}\textbf{T})),
\label{equ:LF}
\end{equation}
where $\textbf{L}$ is the Lagrangean multiplier matrix and $tr(\cdot)$ is the trace of a matrix. In Eq.(\ref{equ:LF}), $tr(\Psi^{\top}\Psi)$ can be rewritten as:
\begin{equation}
\begin{aligned}[b]
\hspace{-0.3cm}tr&\{\Psi^{\top}\Psi\} = tr(\widehat{\textbf{G}}^{\top}\widehat{\textbf{G}}) + c^2\cdot tr(\textbf{T}^{\top}\widehat{\textbf{Z}}^{\top}\widehat{\textbf{Z}}\textbf{T}) + p\cdot \textbf{t}\textbf{t}^{\top}\\ &- 2c\cdot tr(\widehat{\textbf{G}}^{\top}\widehat{\textbf{Z}}\textbf{T}) + 2\cdot tr(\widehat{\textbf{G}}^{\top}\textbf{j}^{\top}\textbf{t}) + 2c\cdot tr(\textbf{T}^{\top}\widehat{\textbf{Z}}^{\top}\textbf{j}^{\top}\textbf{t}),
\label{equ:LF2}
\end{aligned}
\end{equation}
where $p = \textbf{j}\textbf{j}^{\top} = \mu$ is a scalar representing the total rows in the matrices. To infer $\textbf{t}$, we set the derivative of Eq. \ref{equ:LF} w.r.t. $\textbf{t}$ to 0:
\begin{equation}
\frac{\partial \textbf{F}}{\partial \textbf{t}} = 2p\cdot \textbf{t}-2\cdot \widehat{\textbf{G}}^{\top}\textbf{j}^{\top} + 2c\cdot \textbf{T}^{\top}\widehat{\textbf{Z}}^{\top}\textbf{j}^{\top} = 0, \label{equ:DLF2}
\end{equation}
which leads to:
\begin{equation}
\textbf{t} = \frac{(\widehat{\textbf{G}}-c\cdot \widehat{\textbf{Z}}\textbf{T})^{\top}\textbf{j}^{\top}}{p}.
\label{equ:Tra}
\end{equation}

To estimate $\textbf{T}$, we take the derivate of Eq. \ref{equ:LF} w.r.t. $\textbf{T}$:
\begin{equation}
\frac{\partial \textbf{F}}{\partial \textbf{T}} = 2c^2\cdot \widehat{\textbf{Z}}^{\top}\widehat{\textbf{Z}}\textbf{T} -2c\cdot \widehat{\textbf{Z}}^{\top}\widehat{\textbf{G}} +2c\cdot \widehat{\textbf{Z}}^{\top}\textbf{j}^{\top}\textbf{t} + \textbf{T}(\textbf{L}+\textbf{L}^{\top}),
\label{equ:DLF1}
\end{equation}
from which we can see that $\widehat{\textbf{Z}}^{\top}\widehat{\textbf{Z}}$ and  $\textbf{L}^{\top}\textbf{\textbf{L}}$ are both symmetric matrices. After multiplying $\textbf{T}^{\top}$ with Eq.\eqref{equ:DLF1}, we can easily conclude that the term $\textbf{T}^{\top}\widehat{\textbf{Z}}^{\top}\widehat{\textbf{G}}-\textbf{T}^{\top}\widehat{\textbf{Z}}^{\top}\textbf{j}^{\top}\textbf{t}$ is also symmetric. 

Hence, considering Eq. \ref{equ:Tra}, we can formulate a symmetric matrix $\textbf{S}\in \mathbb{R}^{d\times d}$ as:
\begin{equation}
\begin{split}
\textbf{S} &= \textbf{T}^{\top}\widehat{\textbf{Z}}^{\top}\widehat{\textbf{G}}- \textbf{T}^{\top}\widehat{\textbf{Z}}^{\top}\textbf{j}^{\top}\textbf{t}\\
& = \widehat{\textbf{Z}}^{\top}\Big{(}\mathbf{I}-\frac{\textbf{j}^{\top}\textbf{t}}{p}\Big{)},
\end{split}
\label{equ:DLF4}
\end{equation}
here $\mathbf{I}$ represents an identity matrix. To compute \textbf{T}, we utilize singular value decomposition $svd(\cdot)$ to achieve the following equilibrium:
\begin{equation}
\begin{aligned}
svd(\textbf{S}\textbf{S}^{\top}) &= \textbf{T} \cdot svd(\textbf{S}^{\top}\textbf{S})\textbf{T}^{\top}\\
s.t. \,\,\,\, \textbf{V}\textbf{D}_S\textbf{V} &= \textbf{T}\cdot \textbf{W}\textbf{D}_S\textbf{W}^{\top}\textbf{T}^{\top}
\end{aligned}
\end{equation}
where $\textbf{V}$ and $\textbf{W}$ represents orthonormal eigenvector matrices, and $\textbf{D}_S$ represents a diagonal eigenvalue matrix. By solving Eq. \ref{equ:DLF4}, we have:
\begin{equation}
\textbf{T} = \textbf{VW}^{\top}.
\end{equation}

Finally, by making use of Eq.\eqref{equ:DLF2} and Eq.\eqref{equ:Tra}, we compute the scaling factor as follows:
\begin{equation}
c = \frac{tr\Big{(} \textbf{T}^{\top}\widehat{\textbf{Z}}^{\top}\big{(}\textbf{I}-\frac{\textbf{j}^{\top}\textbf{t}}{p}\big{)}\widehat{\textbf{G}}\Big{)}}
{tr\Big{(}\widehat{\textbf{Z}}^{\top}\big{(}\textbf{I}-\frac{\textbf{j}^{\top}\textbf{t}}{p}\big{)}\widehat{\textbf{Z}}\Big{)}}.
\end{equation}

To learn the mapping function for aligning all node embeddings, the above learning process for $\mathcal{M}_{\mathcal{G}_n}(\cdot)$ is repeated $k$ times for all subnetworks. The nodes having $|\mathbb{R}|$ contextual representations in $j$ different subnetworks, we first align them w.r.t. their partition and anchor graph. Then these aligned embeddings $\textbf{z}_a^1,\textbf{z}_b^1,\textbf{z}_c^1 ...$ are averaged to form the final node embedding:
\begin{equation}
	\textbf{z} = \frac{\sum_{j}^{|\mathcal{R}|} \textbf{z}_j}{|\mathcal{R}|},
	\label{eq_average}
\end{equation}
which comprehensively combines each node's high-order information and helps with generalization, using multi-context aggregation.
\vspace{-4mm}

\section{Experiments}
\label{sec:Experiments}
In the experimentation section, we extensively evaluate the node embedding quality and the efficiency of DeHIN. To quantify our evaluation criteria, we adopt node classification and link prediction tasks, and evaluate the efficiency of DeHIN against large-scale graph embedding frameworks. Specifically, research questions (RQs) we aim  investigate are:
\begin{itemize}
	\item[RQ1:] How effective is the partitioning mechanism of DeHIN?
	\item[RQ2:] While handling large-scale networks, how is the embedding quality of DeHIN for the node classification task?
	\item[RQ3:] How good are the embeddings computed by DeHIN for predict unseen links?
	\item[RQ4:] How efficient is DeHIN when embedding large-scale networks as compared to distributed network embedding frameworks?
	Ablation Study 
	\item[RQ5:] How effective is DeHIN after ablating its multi-context alignment mechanism?
\end{itemize}

\subsection{Baseline Methods}
To show the performance of DeHIN, we compare it with several strong baselines that are designed for HNE task.
\begin{itemize}
	\item \textbf{metapath2vec} \cite{dong2017metapath2vec}: In-order to capture the heterogeneity, metapath2vec utilizes the node paths generated through guided random walks. These path specifically captures the node context.
	\item \textbf{PTE} \cite{tang2015pte}: This method decomposes a heterogeneous network into number of bipartite networks, each representing a relationship type. Its objective function computes the sum  of log-likelihood over all bipartite networks.
	\item \textbf{HIN2Vec} \cite{fu2017hin2vec}: HIN2Vec is a neural network model that is also based on metapaths. Given a set of relationships specified by meta-paths, HIN2Vec is able to learn expressive node representations from heterogeneous networks.
	\item \textbf{HEER} \cite{shi2018easing}: This method extends PTE by computing the closeness between types of relationships.
	\item \textbf{R-GCN} \cite{schlichtkrull2018modeling}: R-GCN aggregates neighbours having same edge types and considers edge heterogeneity by learning multiple convolution 
	matrices.
	\item \textbf{HAN} \cite{wang2019heterogeneous}: HAN uses attention mechanism that learns wights of node neighbourhood, sampled using meta-path generation mechanism. 
	\item \textbf{MAGNN} \cite{fu2020magnn}: Embeddings are learned in MAGNN using meta-path based neighborhood and edge type information to parametrizes a transformer-based self attention architecture.
	\item \textbf{HGT} \cite{hu2020heterogeneous}: Transformer-based self-attention model is used in HGT, parametrized by edge type.
	\item \textbf{TransE} \cite{bordes2013translating}:	This method models relationship among nodes by interpreting them as translations operating on the node embeddings.
	\item \textbf{DistMult} \cite{yang2014embedding}: DistMult develops a simple bilinear model by generalizing NTN \cite{socher2013reasoning} and TransE, for learning low dimensional vector representation of network nodes.
	\item \textbf{ComplEx} \cite{trouillon2016complex}:	ComplEx approach the link prediction problem using latent factorization to compute complex valued embeddings. 
	\item \textbf{ConvE} \cite{dettmers2018convolutional}:	ConvE consists of a single convolution layer, projection layer and an inner product layer, specifically designed for the link prediction in knowledge graphs.
	
The embedding dimensions of all baselines is set to 32, and all other hyperparameters are tuned following the original papers. Implementations of these methods in python package is provided by \cite{yang2020heterogeneous}\footnote{https://github.com/yangji9181/HNE}.

\end{itemize}
The above methods are, however, centralized models. These methods incur memory constraints and are time-consuming on a single computing unit. In order to show the efficiency of DeHIN, we also compare it with the following distributed network embedding frameworks:
\begin{itemize}	
	\item \textbf{DeLNE} \cite{imran2020decentralized}: DeLNE adopts data parallelism approach to embed large-scale networks. As it is originally designed for homogeneous graphs, we implement DeLNE on heterogeneous by treating all node and edge types uniformly.
	\item \textbf{PBG} \cite{pbg}: PyTorch-BigGraph  is a distributed system that trains on an input graph by ingesting its list of edges and their relation types. It outputs an embedding for each node by placing linked nodes close in the vector space.
	\item \textbf{DDHH}: DDHH is a large-scale decentralized HIN embedding framework, we presented in \cite{imran2020DDHH}. DDHH adopts centralized partitioning strategy and uses orthogonal alignment mechanism, while only considering the first instance of the node received at the server.
	\item \textbf{DeHIN-Basic}: In order to study the effectiveness of our alignment mechanism and ablation study, we remove alignment settings of DeHIN to construct DeHIN-Basic.
\end{itemize}

\begin{table}[!t]
	\centering
	\setlength{\tabcolsep}{2pt}
	\caption{Statistics of four networks. Connected components (CC) are computed after dividing each network into multiple hypergraphs.}
	\vspace{-0.4cm}
	\begin{tabular}{|l|cccc|}
		\hline
		\textbf{Statistics} & \textbf{Yelp}       & \textbf{PubMed}  & \textbf{DBLP}        & \textbf{Freebase}   \\ \hline
		\#Nodes          & 82,465     & 63,109  & 1,989,077   & 12,164,758 \\
		\#Node Type      & 4          & 4       & 4           & 8          \\
		\#Edges          & 30,542,675 & 244,986 & 275,940,913 & 62,982,566 \\
		\#Edges Type  & 4          & 10      & 6           & 36         \\
		\#Features       & N/A        & 200     & 300         & N/A        \\
		\#Label Type     & 16         & 8       & 13          & 8          \\
		Total CC & 81      & 9,028     & 65,167         & 1,069,314     \\ 
		Largest CC Size  & 82,417      & 24,429     & 1,766,444         & 1,561,926     \\
		Average Nodes per CC  & 7,417      & 454     & 618         & 47,190     \\  
		\hline
		\#DeHIN Partitions  & 4      & 10     & 23         & 339     \\ \hline
	\end{tabular}
	\label{dataset}
	\vspace{-2 mm}
\end{table}

\subsection{Datasets}\label{sec_dataset}
In our experiments, we use four real large-scale heterogeneous networks extracted and labeled by \cite{yang2020heterogeneous}, which are biomedical library (PubMed), business (Freebase), e-commerce network (YELP) and academic networks (DBLP). Data statistics are provided in Table \ref{dataset}. The characteristics of each dataset are as follows:
\begin{itemize}
	\item\textbf{PubMed}\footnote{https://www.nlm.nih.gov/databases/download}: PubMed is a biomedical literature network of genes, diseases, chemical and species. word2vec is used to extract embeddings from PubMed papers to get 200 features and are aggregated with associated nodes. A small portion of diseases are manually labeled into eight categories.  Each labeled disease has only one label.
	\item \textbf{YELP}\footnote{https://www.yelp.com/dataset}: It is constructed using business, users, locations and reviews from YELP network. A few thousand businesses are manually labeled into sixteen categories. Each labeled business has one or multiple labels.
	\item \textbf{DBLP}\footnote{https://dblp.org/xml/release/}: DBLP dataset consists of a network of authors, papers, venues and parsed phrases. Each paper has 300 features, computed using word2vec. Author and venue features are the aggregation of their corresponding paper embeddings. Labels represent authors' research fields, that were manually annotated. Each labeled author has only one label.
	\item \textbf{Freebase}\footnote{https://developers.google.com/freebase/data}: Freebase network is constructed by linking books, films, music, sports, people, locations, organizations, and businesses. Nodes are not associated with any features. Each label represents eight different genres of literature. Each labeled book has only one label.
\end{itemize}

\subsection{Implementation Details}
In our implementation largely adopted two platforms, Python for processing and Oracle of data storage. We set the minimum partition size to $m$ to 10,000 and the upper bound $m'$ is set to 40,000, to balance the density and scale of the data. The bounds are determined after numerous partition and training runs on each dataset. Table \ref{dataset} shows the number of partitions generated for each dataset. Our alignment mechanism further supports integrating high-order relationship information for each node.  For DeHIN workers, we set the embedding dimension $d$ to 32, the number of epochs for each worker model to 100 (early stopping enabled) with a learning rate of 0.01. Due to the availability of maximum of 80 cores, we run workers in batches for training and record the total time as time taken by the slowest worker plus the alignment time. To maintain efficiency, we run 20 worker models simultaneously with each worker utilizing 4 CPU cores at a time.

\begin{figure}[!t]
	\centering
	\includegraphics[width=\linewidth]{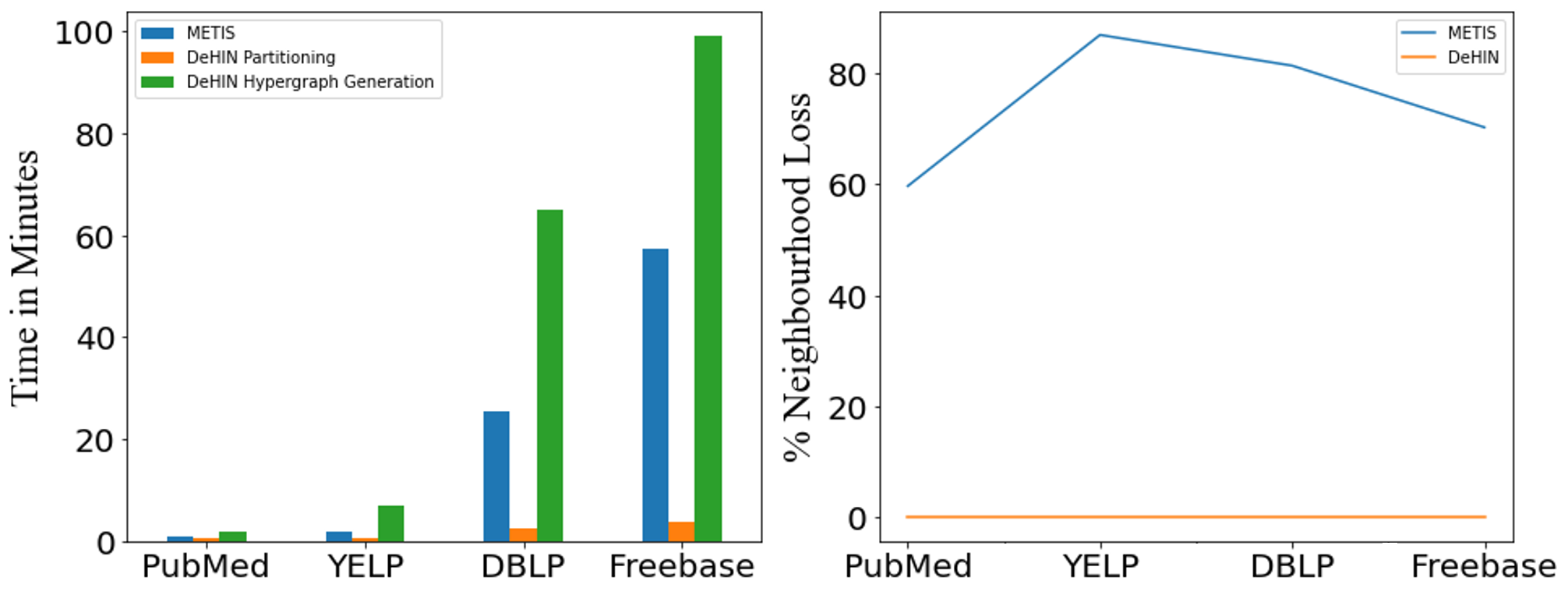}
	\vspace{-0.2cm}
	\caption{ \textbf{A}) Partition time of METIS (blue) and DeHIN (orange), as well as the hypergraph generation time of DeHIN (green) in minutes. \textbf{B}) Partition quality, measured by average neighbourhood loss (\%).
	}
\vspace{-0.2 mm}
	\label{partition_time}
\end{figure}

\subsection{Partition Effectiveness and Efficiency (RQ1)}
To show the effectiveness of our distributed HIN partition module, we compare it against METIS \cite{karypis1999multilevel}, which is widely adopted for graph partitioning task because of its efficiency and partition quality. In our work, we keep the partition size equal to the maximum value between lower-bound and largest connected component. To further evaluate the partition quality of both methods, we compare the two methods w.r.t. time efficiency in minute Fig. \ref{partition_time}\textbf{A} and percentage of average neighbourhood loss for each node. We compare the average loss of each node's neighbourhood in partitioned networks against average neighbourhood of node in the full network which is calculated via,
\begin{equation}
	\%S_{Avg\ Loss} = \frac{\sum\limits_{j=1}^{k}\sum\limits_{r=1}^{|\mathcal{R}|}\sum\limits_{i=1}^{|V^j|}{(S_i^r-S_i^{r,k})}}{\sum\limits_{r=1}^{|\mathcal{R}|}\sum\limits_{i=1}^{|V|}S_i^r-|V|}\times 100
	\label{average}
\end{equation}
where, $S_i$ is the first-order neighbourhood of node $n_i^r$ in context $r$, while $S_i^{k}$ is the first-order neighbourhood of $n_i^r$ in $k^{th}$ partition. Although, hyperedge generation is done in parallel for each relationship type, yet it is costly for networks having long CC. For partition task DeHIN outperforms METIS (Fig. \ref{partition_time}\textbf{A}). METIS produces partitions with high clustering coefficient but fails to preserve adequate neighbourhood structure (Fig. \ref{partition_time}\textbf{B}), as it cuts meaningful edges that are required to learn network`s structural properties. Rather than cutting edges, DeHIN samples hyperedges, each with holding a complete connected neighbourhood within $r$, hence effectively preserving a node's structural properties.  This also affects the final embedding alignment process, where we try to orthogonally align different contexts of a node that are spread out.

\begin{figure}[h]
	\centering
	\includegraphics[width=8.1cm]{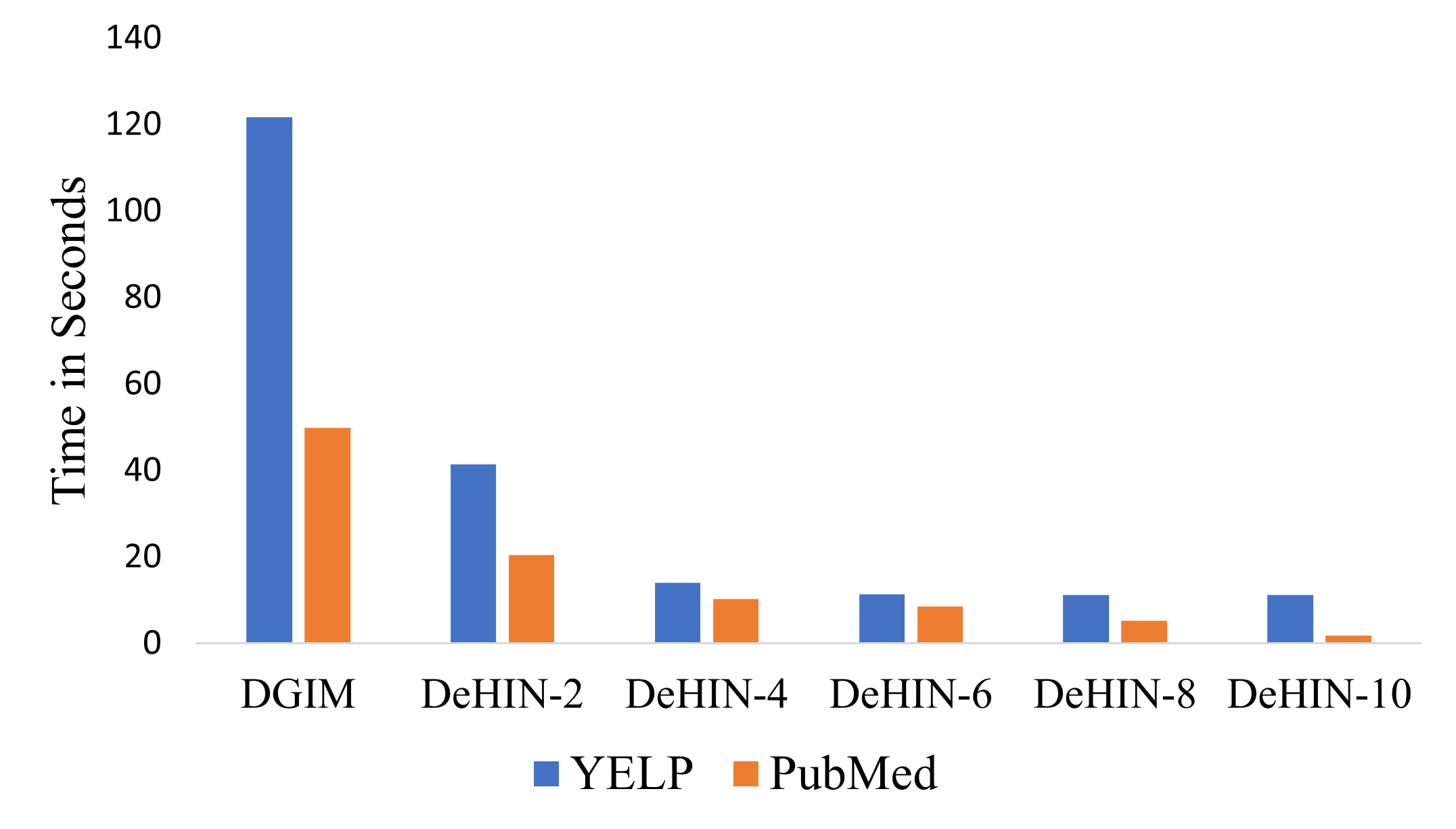}
	\vspace{-3.5 mm}
	\caption{ Comparision of learning time between centralized setting of DeHIN (i.e., DGIM model) and DeHIN with 2,4,6,8 and 10 workers.
	}
	
	\label{central_time}
\end{figure}

In Fig. \ref{central_time} we also compare the running time of a centralized DeHIN (i.e., a DGIM model on a single worker) against DeHIN with 2,4,8 and 10 workers, using two of the smaller datasets i.e., YELP and PubMed. We choose to report the efficiency for these datasets since it is inefficient to compute the mutual information between the node patches and the complete network summary for our large-scale HINs in the case of centralized DeHIN. The time comparison shows that distributed learning reduces the learning time significantly with four or more workers. This is because the DGIM model compares the patch summary of both positive and negative samples against a complete graph summary. For larger HINs, computing a complete graph summary is a very exhaustive process.

\subsection{Node Classification Task (RQ2)}
Using  70\% of labeled nodes, we train a support vector machine (SVM) \cite{fan2008liblinear} classifier to support the node classification task. The rest of the 30\% nodes are used for evaluation purpose. Details of node labeling is explained in Section \ref{sec_dataset}. The reported experimental results in Table \ref{table_classification} are averaged on five runs.  

\begin{table}[]
	\centering
	\setlength{\tabcolsep}{4pt}
	\caption{Results of the node classification task.}
	\begin{tabular}{|l|llllllll}
		\hline
		\multirow{2}{*}{\textbf{Model}} & \multicolumn{8}{c|}{\textbf{Node classification (Macro-F1/Micro-F1)}}                                                                                                      \\ 
		& \multicolumn{2}{c}{\textbf{PubMed}}     & \multicolumn{2}{c}{\textbf{YELP}}        & \multicolumn{2}{c}{\textbf{Freebase}}    & \multicolumn{2}{c|}{\textbf{DBLP}}         \\ \hline
		metapath2vec                    & \multicolumn{2}{l}{12.90/15.51}         & \multicolumn{2}{l}{\textbf{5.16}/23.32}           & \multicolumn{2}{l}{20.55/46.43}          & \multicolumn{2}{l|}{\textbf{43.85}/55.07}           \\
		PTE                             & \multicolumn{2}{l}{09.74/12.27}         & \multicolumn{2}{l}{5.10/23.24}           & \multicolumn{2}{l}{10.25/39.87}          & \multicolumn{2}{l|}{43.34/54.53}           \\
		HIN2Vec                         & \multicolumn{2}{l}{10.93/15.31}         & \multicolumn{2}{l}{5.12/23.25}           & \multicolumn{2}{l}{17.40/41.92}          & \multicolumn{2}{l|}{12.17/25.88}           \\
		HEER                            & \multicolumn{2}{l}{11.73/15.29}         & \multicolumn{2}{l}{5.03/22.92}           & \multicolumn{2}{l}{12.96/37.51}          & \multicolumn{2}{l|}{09.72/27.72}           \\
		R-GCN                           & \multicolumn{2}{l}{10.75/12.73}         & \multicolumn{2}{l}{5.10/23.24}           & \multicolumn{2}{l}{06.89/38.02}          & \multicolumn{2}{l|}{09.38/13.39}           \\
		HAN                             & \multicolumn{2}{l}{09.54/12.18}         & \multicolumn{2}{l}{5.10/23.24}           & \multicolumn{2}{l}{06.90/38.01}          & \multicolumn{2}{l|}{07.91/16.98}           \\
		MAGNN                           & \multicolumn{2}{l}{10.30/12.60}         & \multicolumn{2}{l}{5.10/23.24}           & \multicolumn{2}{l}{06.89/38.02}          & \multicolumn{2}{l|}{06.74/10.35}           \\
		HGT                             & \multicolumn{2}{l}{11.24/18.72}         & \multicolumn{2}{l}{5.07/23.12}           & \multicolumn{2}{l}{23.06/46.51}          & \multicolumn{2}{l|}{15.17/32.05}           \\
		TransE                          & \multicolumn{2}{l}{11.40/15.16}         & \multicolumn{2}{l}{5.05/23.03}           & \multicolumn{2}{l}{31.83/52.04}          & \multicolumn{2}{l|}{22.76/37.18}           \\
		DistMult                        & \multicolumn{2}{l}{11.27/15.79}         & \multicolumn{2}{l}{5.04/23.00}           & \multicolumn{2}{l}{23.82/45.50}          & \multicolumn{2}{l|}{11.42/25.07}           \\
		ComplEx                         & \multicolumn{2}{l}{09.84/18.51}         & \multicolumn{2}{l}{5.05/23.03}           & \multicolumn{2}{l}{35.26/52.03}          & \multicolumn{2}{l|}{20.48/37.34}           \\
		ConvE                           & \multicolumn{2}{l}{\textbf{13.00}/14.49}         & \multicolumn{2}{l}{5.09/23.02}           & \multicolumn{2}{l}{24.57/47.61}          & \multicolumn{2}{l|}{12.42/26.42}           \\ \hline
		\textbf{DeHIN}                 & \multicolumn{2}{l}{{12.07}/\textbf{24.30}} & \multicolumn{2}{c}{5.02/\textbf{27.02}} & \multicolumn{2}{l}{\textbf{42.80/56.56}} & \multicolumn{2}{l|}{28.49/\textbf{55.11}}\\ \hline
	\end{tabular}
\label{table_classification}
\end{table}

For the task of node classification, DeHIN produces comparable results with other state-of-the-art methods. Due to the presence of increased relationship information (36 links) in Freebase network, alignment mechanism of DeHIN is more able to capture high-order node relationships and hence improved classification accuracy as compared with the rest. Low Macro-F1 score in the case of PubMed, Yelp and DBLP networks is mainly due to very high class imbalance in the labeled data. Macro-F1 is a straightforward averaged result for all classes. Though metapath2vec performs well on several minority classes and thus achieves a higher Macro-F1, it does not necessarily mean that its has more correct predictions than DeHIN on all test cases.  When the classes are highly imbalance, Macro-F1 is less representative for a model's performance. So in DBLP, Micro-F1 is a more reliable metric and it achieves the best performance among all baselines.
Comparable Micro-F1 score for DeHIN suggest that the learned embeddings of the labeled nodes, across all datasets, can distinguish positive predictions. 

\subsection{Link Prediction Task (RQ3)}

For the link prediction task, Hadamard function is used to construct feature vectors for node pairs. 20\% of the links are hidden before the training while two-class LinearSVC is trained on rest of the 80\% training links. The resulted AUC (area under the ROC curve) is the resultant of 5 iterations.

DeHIN distinguishes true links from false with comparable accuracy against centralized models in Table \ref{link_pred}. However, for the two datasets (YELP and DBLP) having only 4 relationship types and very high link-type imbalance i.e. YELP having 90\% \textit{phrase-context-phrase} link among total link types and DBLP with 89\% Author Year Author link as majority link type, metapath2vec (80.52/80.31) and HEER (67.2/61.71) have slightly better performance. This is because DeHIN captures high-order information in embeddings by capturing relationship types from node links. For the datasets having higher link type diversity, DeHIN is better at predicting true from false links and vice versa.

\begin{table}[!t]
	\centering
	\caption{Results of the link prediction task.}
	\begin{tabular}{|l|l|l|c|l|c|l|c|l|}
		\hline
		\multirow{2}{*}{Model} & \multicolumn{8}{c|}{\textbf{Link Prediction Accuracy (AUC)}}                                                                        \\ 
		& \multicolumn{2}{c}{\textbf{PubMed}} & \multicolumn{2}{c}{\textbf{YELP}}  & \multicolumn{2}{c}{\textbf{Freebase}} & \multicolumn{2}{c|}{\textbf{DBLP}}  \\ \hline
		metapath2vec           & \multicolumn{2}{l}{69.38}  & \multicolumn{2}{c}{\textbf{80.52}} & \multicolumn{2}{c}{56.14}    & \multicolumn{2}{c|}{65.26} \\ 
		PTE                    & \multicolumn{2}{l}{70.36}  & \multicolumn{2}{c}{50.32} & \multicolumn{2}{c}{57.89}    & \multicolumn{2}{c|}{57.72} \\ 
		HIN2Vec                & \multicolumn{2}{l}{69.68}  & \multicolumn{2}{c}{51.64} & \multicolumn{2}{c}{58.11}    & \multicolumn{2}{c|}{53.29} \\ 
		HEER                   & \multicolumn{2}{l}{68.31}  & \multicolumn{2}{c}{76.1}  & \multicolumn{2}{c}{55.8}     & \multicolumn{2}{c|}{\textbf{67.2}}  \\ 
		R-GCN                  & \multicolumn{2}{l}{69.06}  & \multicolumn{2}{c}{73.72} & \multicolumn{2}{c}{55.78}    & \multicolumn{2}{c|}{53}    \\ 
		HAN                    & \multicolumn{2}{l}{63.33}  & \multicolumn{2}{c}{72.17} & \multicolumn{2}{c}{50.18}    & \multicolumn{2}{c|}{50.5}  \\ 
		MAGNN                  & \multicolumn{2}{l}{65.85}  & \multicolumn{2}{c}{N/A}   & \multicolumn{2}{c}{51.5}     & \multicolumn{2}{c|}{50.24} \\ 
		HGT                    & \multicolumn{2}{l}{61.11}  & \multicolumn{2}{c}{50.03} & \multicolumn{2}{c}{50.12}    & \multicolumn{2}{c|}{50.1}  \\ 
		TransE                 & \multicolumn{2}{l}{73.00}     & \multicolumn{2}{c}{79}    & \multicolumn{2}{c}{55.68}    & \multicolumn{2}{c|}{59.98} \\ 
		DistMult               & \multicolumn{2}{l}{67.95}  & \multicolumn{2}{c}{69.13} & \multicolumn{2}{c}{52.84}    & \multicolumn{2}{c|}{63.53} \\ 
		ComplEx                & \multicolumn{2}{l}{70.61}  & \multicolumn{2}{c}{80.28} & \multicolumn{2}{c}{54.91}    & \multicolumn{2}{c|}{52.87} \\ 
		ConvE                  & \multicolumn{2}{l}{{75.96}}  & \multicolumn{2}{c}{80.11} & \multicolumn{2}{c}{60.43}    & \multicolumn{2}{c|}{65.92} \\  \hline
		\textbf{DeHIN }                & \multicolumn{2}{l}{\textbf{78.74}}       & \multicolumn{2}{c}{80.31} & \multicolumn{2}{c}{\textbf{64.51}}         & \multicolumn{2}{c|}{61.71} \\ \hline
	\end{tabular}
\label{link_pred}
\end{table}

\begin{table}[t]
	\caption{Hyperparameter settings for PBG, DeLNE and DeHIN. Number of workers are given in the order of PubMed/YELP/DBLP/Freebase and LR represents the learning rate of each worker.}
	\begin{tabular}{|l|ccccc|}
		\hline
		\textbf{Models} & \textbf{RAM} & \textbf{\#CPUs} & \textbf{\#Workers} & \textbf{LR} & \textbf{\#Epochs} \\ \hline
		\textbf{PBG}    & 1008 GB                          & 80                               & N/A                                  & 0.1                              & 100                                  \\
		\textbf{DeLNE}  & 1008 GB                          & 80                               & 4/12/150/500                         & 0.1                              & 100                                  \\
		\textbf{DeHIN}  & 1008 GB                          & 80                               & 10/5/98/343                         & 0.1                              & 100                                  \\ \hline
	\end{tabular}
\label{hyper}
\end{table}

\subsection{Efficiency Analysis (RQ4)}
All of the centralized models are designed to improve node classification and link prediction accuracies. For large-scale HINs, these models take days to converge and learn embeddings. In order to test the efficiency of DeHIN, we compare it with models that support parallel computations. The Hyperparameter settings for each framework are given in Table \ref{hyper}. 

\begin{table*}[h]
	\centering
	\setlength{\tabcolsep}{3pt}
	\caption{Processing time (P.t), model learning time (L.t), total running time (T.t) of PBG, DeLNE and DeHIN in minutes. The Micro-F1 scores of node classification are also reported.}
	\vspace{-0.4cm}
	\begin{tabular}{|l|ccc|c|ccc|c|cccc|ccc|c|}
		\hline
		& \multicolumn{4}{c|}{\textbf{PubMed}}                   & \multicolumn{4}{c|}{\textbf{YELP}}                      & \multicolumn{4}{c|}{\textbf{DBLP}}                                            & \multicolumn{4}{c|}{\textbf{Freebase}}                  \\ \cline{2-17} 
		\textbf{Models} & P.t  & L.t  & T.t      & Micro-F1       & P.t  & L.t  & T.t       & Micro-F1       & P.t  & L.t  & \multicolumn{1}{c|}{T.t }      & Micro-F1       & P.t  & L.t  & T.t       & Micro-F1       \\ \hline
		\textbf{PBG}    & 2.54      & 3.62      & 6.16          & 15.02          & 7.62      & 367.54    & 375.16         & 22.92          & 67.43     & 4081.59   & \multicolumn{1}{c|}{4149.02}        & 37.08          & 103.88    & 5632.47   & 5736.35        & 51.79          \\
		\textbf{DeLNE}  & 1.20      & 0.88      & 2.08          & 11.20          & 23.59     & 139.65    & 163.24         & 13.43          & 36.10     & 236.25    & \multicolumn{1}{c|}{272.35}         & 21.43          & 87.21     & 423.05    & 510.26         & 23.45          \\

		\textbf{DDHH}  & 3.20      & 1.23      & 4.43          & 21.23          & 25.11     & 1.23    & 26.34         & 31.45          & 39.78     & 52.68    & \multicolumn{1}{c|}{92.46}         & 0.4222          & 93.91     & 61.32    & 155.23         & 52.42          \\

		\textbf{DeHIN-Basic}  & 0.52      & 1.19   & 1.71 & 14.19 & 3.47      & 10.45     & 13.92 & 19.34 & 8.80       & 51.46     & \multicolumn{1}{c|}{60.34} & 28.96 & 10.40     & 60.01     & 70.41 & 43.16 \\
		
		\textbf{DeHIN}  & 0.52      & 1.23      & 1.75 & \textbf{24.30} & 3.47      & 10.50     & 13.97 & \textbf{27.02} & 8.80       & 52.63     & \multicolumn{1}{c|}{61.43} & \textbf{55.11} & 10.40     & 61.32     & 73.72 & \textbf{58.35} \\ \hline
	\end{tabular}
	\vspace{-0.2cm}
\label{running_time}
\end{table*}

As depicted in Table \ref{running_time}, DeHIN outperforms both PBG and DeLNE in its time consumption for training and accuracy for multi-node classification task. The significantly lower time consumption, against PBG and DeLNE comes from three aspects. Firstly, it pre-processes the data and generates partitions by utilizing decentralized computations. Secondly, owing to our distributed architecture with no requirement on model parameter sharing across workers, DeHIN bypasses the potentially heavy communication overhead. Thirdly, since our distributed subnetwork embedding model can directly aggregate patch information for each node, it is computationally efficient. 
Though the node classification accuracy of PBG is comparable with DeHIN, having a large number of relationships makes it harder for PBG to maintain parallelism, inuring heavy communication overhead. 

\subsection{Ablation Study (RQ5)}
Table \ref{running_time} also depicts the ablation study of DeHIN against its basic settings i.e., DeHIN-basic and DDHH (our previous work). DeHIN-basic has no alignment mechanism, while DDHH completely neglects the broader semantic context of a common node by keeping only its fist instance received at the server. We can see in Table \ref{running_time} that both, the orthogonal procrustes alignment and multi-context aggregation support heavenly in efficiently preserving high-order context information, as DeHIN outperforms all baselines for node classification task.

\section{Conclusion}
\label{sec:conclusion}
Learning node embeddings on HIN that consists of billions of nodes and links, creates a performance bottleneck for existing centralized HNE methods. To boost the HNE task with efficiency and embedding quality, we propose \textit{Decentralized Embedding Framework for Heterogeneous Information Network} (DeHIN) in this paper. In DeHIN, we generate a distributed parallel pipeline that utilizes hypergraphs in-order to infuse parallelization in HNE task. DeHIN innovatively formulates large-scale HINs into hypergraph to support parallel HIN partitioning. It then employs deep information maximization theorem on distributed workers, to locally learn node embeddings for each subnetwork. We also purpose a salable context aware alignment mechanism, that project node embeddings onto a public vector space and aggregates the distributed context. We show in our experimental results the significant improvement of DeHIN against state-of-the-art models, by effectively capturing multi-order context of nodes and its distributive nature.

In this work, we presented an efficient HIN partition mechanism, along with an alignment mechanism that can effectively align unseen node embeddings to a common vector space. We can extend the notion of aligning unseen embedding in future research to find similarities in unseen model parameters. This is especially useful in the case of federated learning, where each worker has personal data rather than shared data. Similarity-based model alignment model aggregation can boost local convergence in the case of federated learning.

\ifCLASSOPTIONcompsoc
\section*{Acknowledgments}
\else
\section*{Acknowledgment}
\fi

This work is supported by Australian Research Council Future Fellowship (Grant No. FT210100624), Discovery Project (Grant No. DP190101985) and  National Natural Science Foundation of China (No. 61972069, 61836007, 61832017).

\ifCLASSOPTIONcaptionsoff
\newpage
\fi

\vskip -15pt plus -1fil

\vskip -15pt plus -1fil
\begin{IEEEbiography}[{\includegraphics[width=1in,height=1.25in,clip,keepaspectratio]{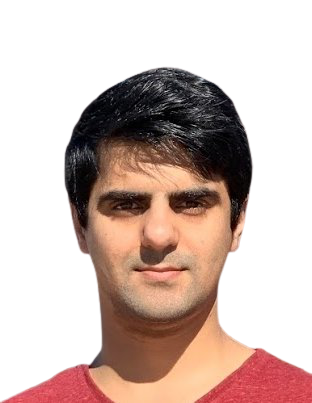}}]{Mubashir Imran}
	Mubashir is currently a computer science Ph.D. student at the School of Information Technology and Electrical Engineering, The University of Queensland. His research interests include data mining, machine learning, and artificial intelligence. More specifically he is currently conducting research in decentralized embedding frameworks and federated recommender systems.
\end{IEEEbiography}
\vskip -15pt plus -1fil

\begin{IEEEbiography}[{\includegraphics[width=1in,height=1.25in,clip,keepaspectratio]{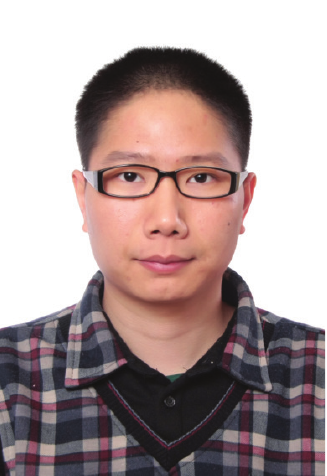}}]{Hongzhi Yin}
received the Ph.D. degree in computer science from Peking University in 2014. He is an Associate Professor with the University of Queensland. 
He received the Australia Research Council Future Fellowship and Discovery Early-Career Researcher Award in 2021 and 2016. He was recognized as Field Leader of Data Mining \& Analysis in The Australian's Research 2020 magazine.
His research interests include recommender system, graph embedding and mining, chatbots, social media analytics and mining, edge machine learning, trustworthy machine learning, decentralized and federated learning, and smart healthcare.

\end{IEEEbiography}
\vskip -15pt plus -1fil
\begin{IEEEbiography}[{\includegraphics[width=1in,height=1.25in,clip,keepaspectratio]{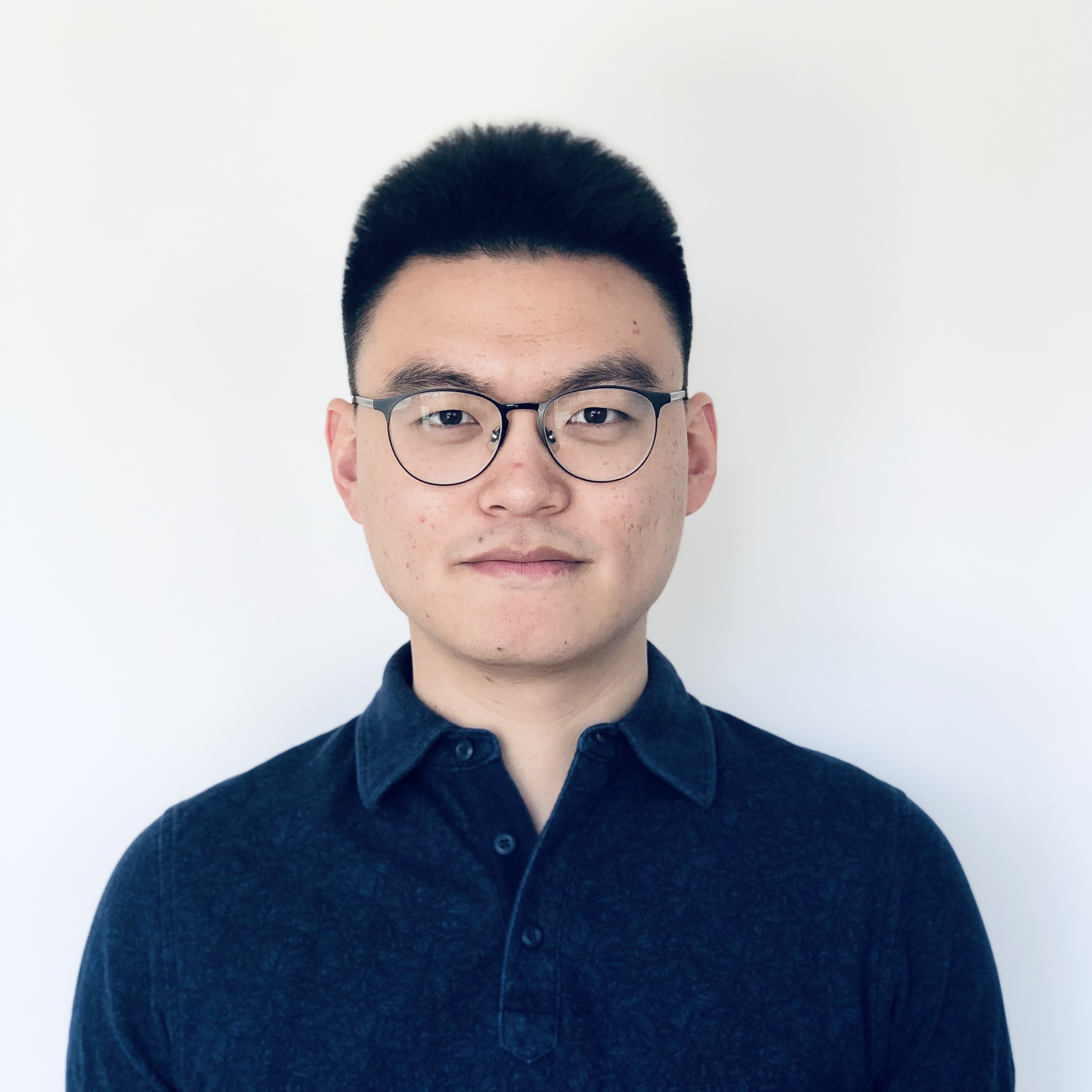}}]{Tong Chen}
received his PhD degree in computer science from The University of Queensland in 2020. He is currently a Lecturer in Business Analytics with Data Science Group, School of ITEE, The University of Queensland. His research interests include data mining, recommender systems, user behavior modelling and predictive analytics. 
\end{IEEEbiography}
\vskip -15pt plus -1fil
\begin{IEEEbiography}[{\includegraphics[width=1in,height=1.25in,clip,keepaspectratio]{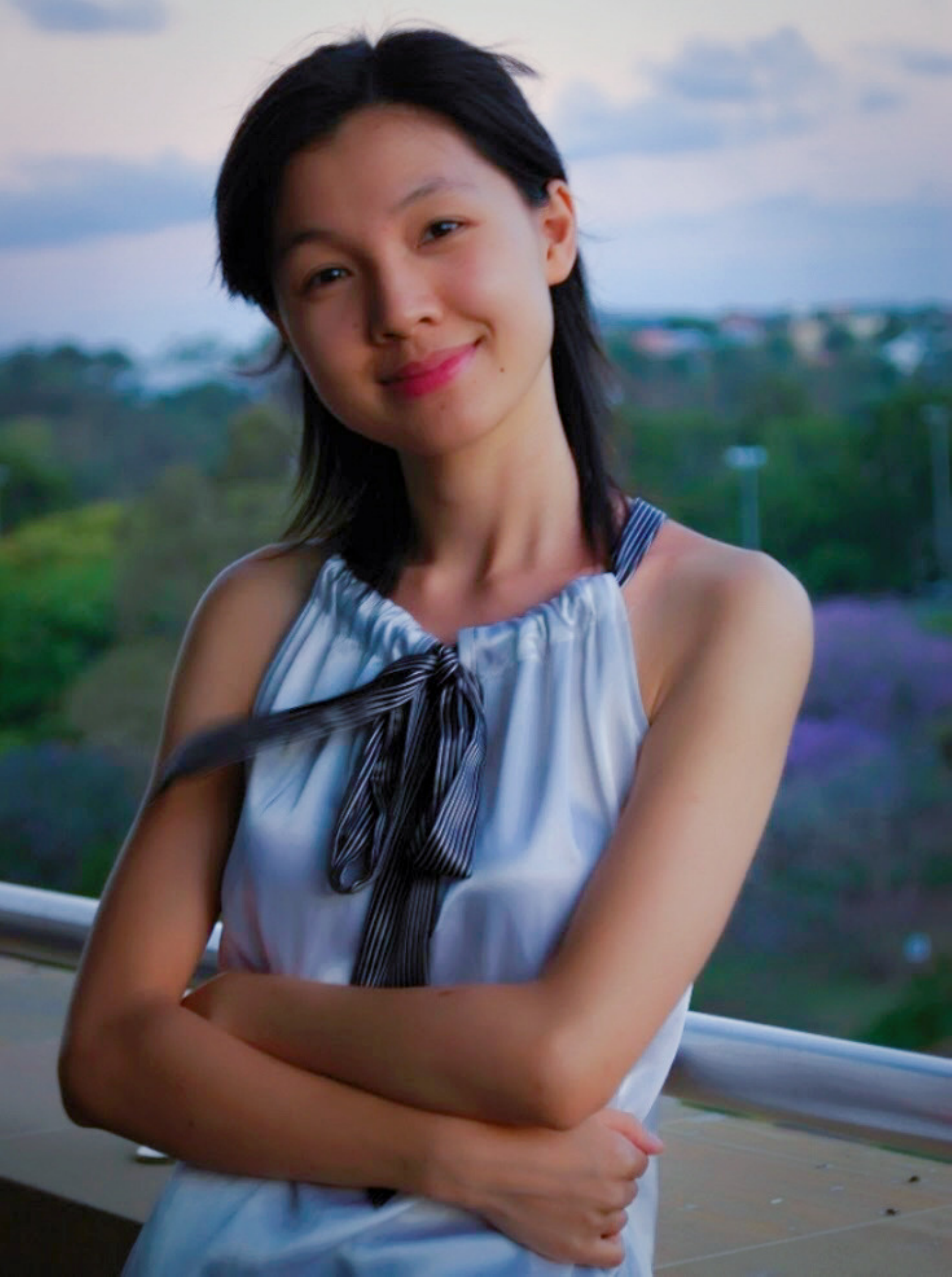}}]{Zi Huang}
received her BSc degree from the Department of Computer Science, Tsinghua University, China, and the PhD degree in computer science from The University of Queensland. She is a professor and ARC Future Fellow in The School of Information Technology and Electrical Engineering, The University of Queensland. Her research interests mainly include multimedia indexing and search, social data analysis, and knowledge discovery.
\end{IEEEbiography}
\vskip -15pt plus -1fil
\begin{IEEEbiography}[{\includegraphics[width=1in,height=1.25in,clip,keepaspectratio]{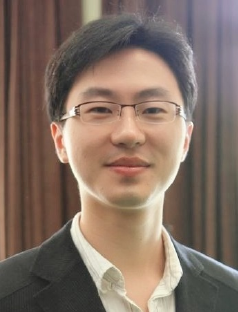}}]{Kai Zheng} is a professor of Computer Sci- ence with University of Electronic Science and Technology of China. He received his PhD degree in Computer Science from The University of Queensland in 2012. He has been working in the area of spatial-temporal databases, uncertain databases, social-media analysis, in-memory computing and blockchain technologies. 
\end{IEEEbiography}

\end{document}